\documentclass[10pt,twocolumn,letterpaper]{article}

\usepackage{wacv}
\usepackage{times}
\usepackage{epsfig}
\usepackage{graphicx}
\usepackage{amsmath}
\usepackage{amssymb}

\usepackage[utf8]{inputenc}
\usepackage[export]{adjustbox}
\usepackage{float}
\usepackage{tabularx, booktabs} 
\usepackage{multirow}
\usepackage{subcaption}
\usepackage{bm}

\graphicspath{{images/}}
\DeclareGraphicsExtensions{.pdf,.jpeg,.jpg,.png}

\newcommand{\Norm}[1]{\left\lVert#1\right\rVert}
\newcommand{\norm}[1]{\left\lvert#1\right\rvert}
\newcommand{\OHR}{\Omega_\text{HR}}
\newcommand{\OLR}{\Omega_\text{LR}}
\newcommand{\R}{\mathbb{R}}
\newcommand{\C}{\text{C}}
\newcommand{\Reg}{\mathcal{R}}
\newcommand{\Rot}{\mathbf{R}}
\newcommand{\Roti}{\mathbf{R}_i}
\newcommand{\T}{\mathbf{T}}
\newcommand{\trans}{\mathbf{t}}
\newcommand{\transi}{\mathbf{t}_i}
\newcommand{\I}{\mathcal{I}}

\newcommand{\zz}{z_0}
\newcommand{\dotp}[1]{\left\langle#1\right\rangle}
\newcommand{\z}{z}
\newcommand{\bw}{\bm{w}}
\newcommand{\bomega}{\bm{\omega}}

\newcommand{\bxi}{\bm{\xi}}
\newcommand{\bxii}{\bm{\xi}_i}
\newcommand{\dx}{\mathrm{d}}
\newcommand{\dA}{\mathrm{d}\mathcal{A}}
\newcommand{\mI}{\bm{I}}
\newcommand{\mIi}{\bm{I}_i}
\newcommand{\mtIic}{\tilde{\bm{I}}_{i,c}}
\newcommand{\mrho}{\bm{\rho}}
\newcommand{\mrhoc}{\bm{\rho}_c}
\newcommand{\p}{\bm{p}}
\newcommand{\pc}{\bm{c}}
\newcommand{\n}{\bm{n}}
\newcommand{\ml}{\bm{l}}
\newcommand{\mli}{\bm{l}_i}
\newcommand{\m}{\bm{m}}
\newcommand{\zi}{z_i}
\newcommand{\epsiloniz}{\varepsilon_{z_i}}

\renewcommand{\epsilon}{\varepsilon}
\renewcommand{\S}{\mathbb{S}}
\renewcommand{\P}{\bm{P}}

\DeclareMathOperator*{\argmin}{arg min}

\DeclareMathOperator*{\mean} {mean}

\usepackage[pagebackref=true,breaklinks=true,letterpaper=true,colorlinks,bookmarks=false]{hyperref}

\wacvfinalcopy 

\def\wacvPaperID{0132} 

\ifwacvfinal\fi
\begin{document}

\title{Inferring Super-Resolution Depth from a Moving Light-Source Enhanced RGB-D Sensor: A Variational Approach}

\author{Lu Sang \qquad Bjoern Haefner \qquad Daniel Cremers\\
Technical University of Munich, Germany \\
{\tt\small {\{lu.sang,bjoern.haefner,cremers\}}@tum.de}
}

\maketitle
\ifwacvfinal\thispagestyle{empty}\fi

\begin{abstract}
A novel approach towards depth map super-resolution using multi-view uncalibrated photometric stereo is presented.
Practically, an LED light source is attached to a commodity RGB-D sensor and is used to capture objects from multiple viewpoints with unknown motion.
This non-static camera-to-object setup is described with a nonconvex variational approach such that no calibration on lighting or camera motion is required due to the formulation of an end-to-end joint optimization problem.
Solving the proposed variational model results in high resolution depth, reflectance and camera pose estimates, as we show on challenging synthetic and real-world datasets.
\end{abstract}
\section{Introduction}
RGB-D sensors are a cheap and easy way to capture RGB and depth images.
Yet, the delivered depth is prone to noise, quantization, missing data and a lower resolution compared to its RGB image.
The Intel RealSense D415 delivers RGB images almost twice as large in resolution compared to its companion depth images.
As the RGB data is substantially better, it seems a good choice to rely on the image intensities to perform depth refinement.
Yet, due to its complexity improving a sensors depth map remains an open and challenging problem in the computer vision community.
It is often tackled by capturing a sequence of RGB-D frames under multiple view-points.
The data, along with the corresponding RGB information can then be used in order to refine the geometric measurements.

A classic technique to estimate geometry with fine scale detail is called Photometric Stereo (PS).
It relies on an image sequence where each frame is captured under different illumination.
Then, PS estimates fine scale details of the geometry, as well as reflectance and possibly lighting properties of the scene.
In classic PS approaches the camera is assumed to be static with respect to the object, while only illumination varies for each frame.

\begin{figure}[!ht]
  \centering
  \setlength\tabcolsep{1pt} 
  \def\arraystretch{0.5} 
  \newcommand{\mywidth}{0.23\textwidth} 
  \newcommand{\mywidthlr}{0.1\textwidth} 

  \newcolumntype{X}{>{\centering\arraybackslash}m{\mywidth}}
  \begin{tabular}{XX}
    \includegraphics[height=0.16\textwidth]{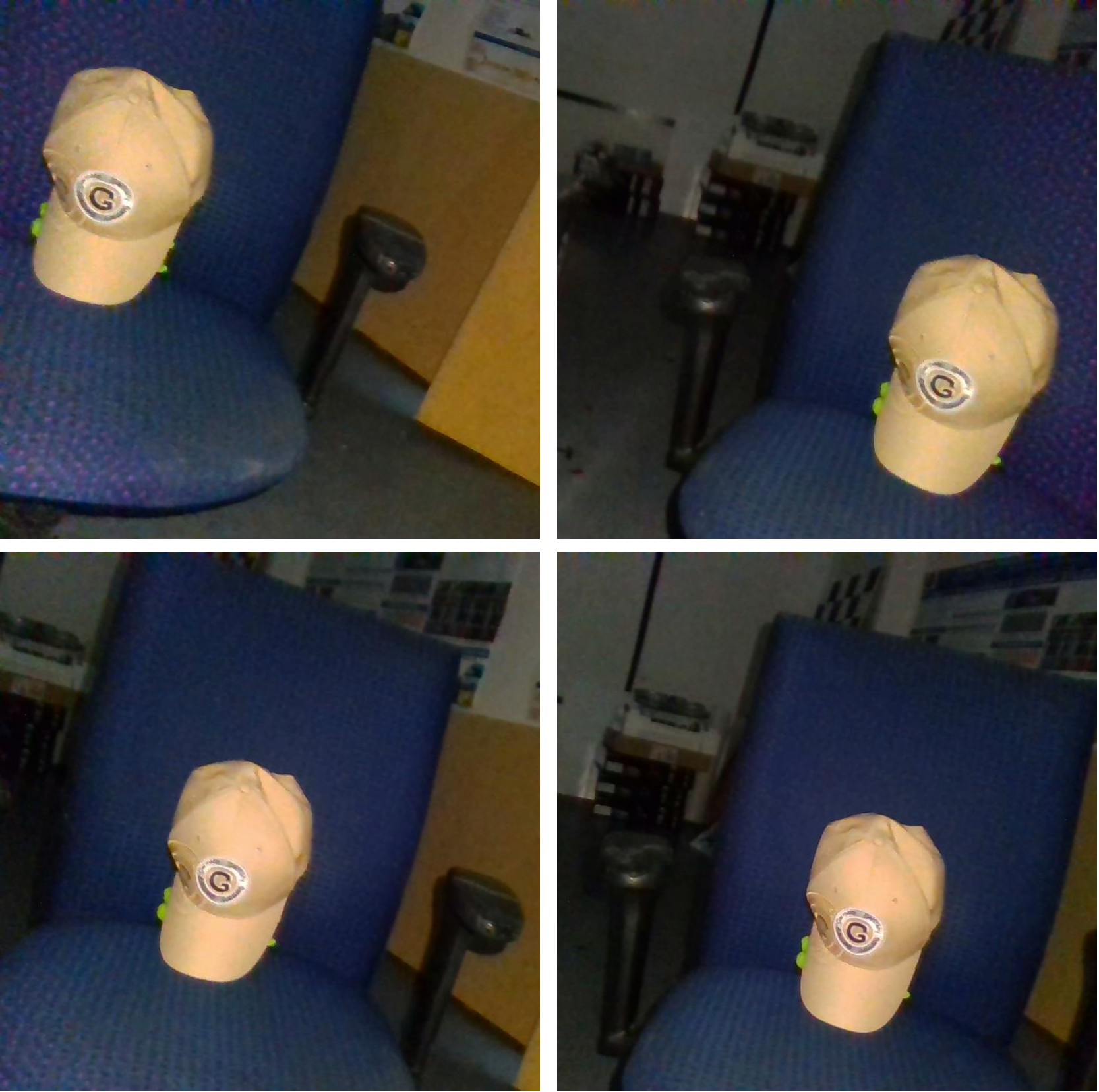}&
    \includegraphics[height=\mywidthlr]{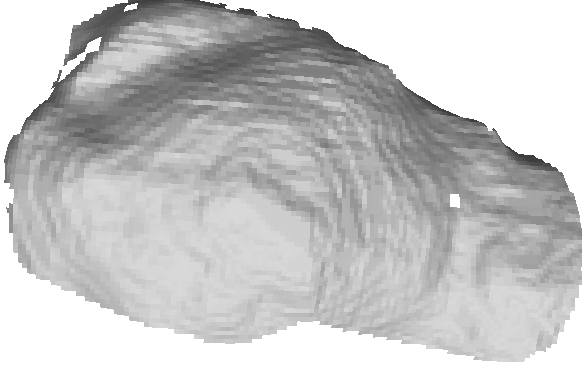}\\
    Selection of Input RGBs & Input Reference Depth\\
    \includegraphics[height=\mywidth, angle=90]{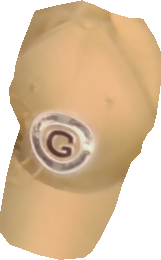}&
    \includegraphics[height=0.15\textwidth]{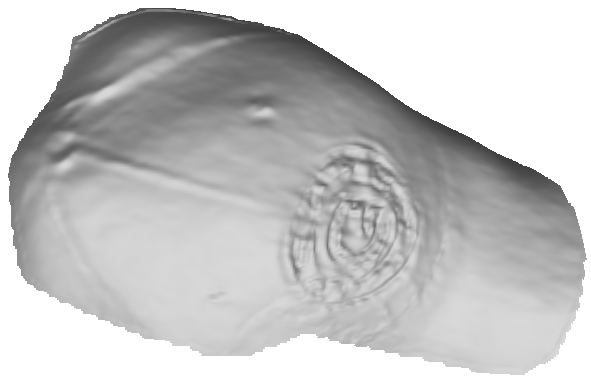}\\
    Estimated Albedo & Estimated Depth
  \end{tabular}
  \caption{We present a novel method that combines depth super-resolution and multi-view uncalibrated photometric stereo.
  Our approach is able to recover a refined high-resolution depth and reflectance estimate based on multiple RGB images and its companion depth map acquired with a moving RGB-D camera and an attached LED light source.}
  \label{fig:teaser}
\end{figure}

For many purposes the classical PS setup is rather impractical, as a moving camera is usually used to capture the object of interest.
We propose to close the gap between photometric stereo and camera motion in order to be able to estimate a high resolution depth map, cf. Figure~\ref{fig:teaser}.
To achieve varying illumination in each frame, we simply attach an LED light source to the RGB-D camera.
No further calibration on the light source is needed, as we put forward an end-to-end variational model which estimates camera motion, depth, reflectance, and lighting in an alternating manner.

The rest of the paper is organized as follows: Section \ref{sec2} reviews related work, while the proposed model is introduced in Section \ref{sec3}.
This is followed by the numerical details in Section \ref{sec4}.
The model is evaluated in Section \ref{sec5} on both, synthetic and real-world datasets.
Eventually, Section \ref{sec6} summarizes and suggests future research directions.
\section{Related Work}\label{sec2}
Let us first give some background information on the two problems of \textit{depth super-resolution} and \textit{photometric stereo}.
\paragraph{Depth Super-Resolution.}
Super-Resolution (SR) is a problem which aims at enhancing the resolution of given imaging data.
The two most common data sources are RGB and depth sensors.
Thus, one commonly differs between two problem statements: image SR~\cite{dong2014,dong2015,goldluecke2014,ledig2017,yang2008,yang2010} which tries to solve SR for RGB images, and depth SR~\cite{kiechle2013,li2012,lu2011,riegler2016} which considers depth maps as the source of data to be enhanced.
The applications of SR range from surveillance~\cite{cristani2004} and medical imaging~\cite{greenspan2008} to multi-view reconstruction~\cite{goldluecke2014} and RGB-D sensing~\cite{maier2015,park2014}.
Our approach is more oriented towards the latter, and we aim to tackle the problem of depth SR which can be largely classified into two categories:
single depth map SR and multi depth map SR.
The former tries to infer a high resolution depth map from a single observation.
One can for instance use an adaption of the non-local means formulation to data typical for depth maps~\cite{huhle2010}.
Another variant is to incorporate the companion RGB image, e.g. to guide a bilateral filtering process and sub-pixel refinement~\cite{yang2007} or to use an anisotropic total generalized variation regularization term in the optimization problem~\cite{ferstl2013}.
Multi-scale guided convolutional networks can complement low-resolution depth features with high resolution RGB features~\cite{takwai2016}, or combinations of classical variational approaches with deep networks were used~\cite{gernot2016}.
Another variant is to train a network end-to-end to directly learn a mapping between low-resolution and high-resolution depth images~\cite{xibin2017}.
Recently, it was shown that taking into account photometric approaches helps solving the problem of depth super-resolution, due to incorporating the physical relationship between intensity and depth~\cite{haefner2019tpami,haefner2018cvpr,lu2013,peng2017}.
Our approach builds upon these ideas to recover a fine scale, geometrically detailed depth image based on a methodology called photometric stereo.
\paragraph{Photometric Stereo.}
Before diving into PS, let us first recall its pioneering work ``Shape-from-Shading''~\cite{horn1989} which aims at recovering shape from a single image by describing the interference between light, geometry and reflectance of the object.
It is impossible to unambiguously infer geometry, even if reflectance and lighting is known.
One way to disambiguate shape-from-shading is to use multiple images instead of a single observation, the clue is to capture each image under different illumination.
This approach is called photometric stereo~\cite{woodham1980} and tries to estimate the objects geometry and possibly its reflectance.
One differs between calibrated PS and uncalibrated PS, where former assumes proper knowledge of the lighting conditions in the scene~\cite{queau2017jmiv}, and the latter, more complicated and ill-posed case, tries to solve PS without lighting information~\cite{basri2007,belhumeur1999}.
Besides the complexity of uncalibrated PS in lighting, it furthermore tends to assume a static camera-to-object relation, i.e. illumination is the only quantity changing in the scene.
This makes PS impractical for applications where an object is captured under camera motion.
Multi-view photometric stereo under unconstrained camera motion and illumination was presented in~\cite{ackermann2014,treuille2004}, yet one needs example objects in order to recover reflectance properties.
Other approaches need additional lighting calibration using a specular sphere during the capturing process and additionally capture multiple images from the same viewpoint under different illumination before moving the camera and repeating this procedure~\cite{zhou2013}.
One can circumvent this by attaching a light source to the camera which then needs to be calibrated beforehand along with the extrinsic parameters of the sensor~\cite{higo2009,simakov2003}.
Placing the object on a turntable and capture it from different viewpoints by rotating it implies different illumination on the object as well, even if lighting is static with respect to the camera~\cite{hernandez2008,zuo2017}.
Although this allows for a non-static camera-to-object setup, the motion of the object relative to the camera remains limited due to the turntable setup.
Structure-from-motion (SfM)~\cite{ullman1979} can be used to estimate odometry and an initial geometry~\cite{zhang2003}, yet SfM relies on the photo consistency assumption which states that the same 3D point in the scene results in the same intensity across images taken from different viewpoints.
When applying PS this assumption is heavily violated, which makes it difficult to apply a SfM approach out-of-the-box.
A similar approach was considered in~\cite{logothetis2018}.
This is realized using LEDs soldered to a circuit board which is attached to a camera.
The camera captures multiple images under different illumination, before repeating this process from different viewpoints.
As the relative position of the LEDs to the camera is fixed and calibrated offline this approach performs calibrated PS.
The advantage of having a sensor based depth estimate from an RGB-D camera seems to resolve the ambiguity arising in the uncalibrated PS case~\cite{peng2017}, which gives rise to use a low-resolution depth map.
Only two approaches use RGB-D cameras in combination   with multi-view PS~\cite{bylow2019,zuo2017}.
The capturing process in~\cite{bylow2019} is closely related to~\cite{logothetis2018,zhou2013}, and~\cite{zuo2017} make use of a turnable.
Thus,~\cite{bylow2019,zuo2017} remain limited to their specific setup.
\\
To the best of our knowledge, we are the first ones who not only combine both, depth super-resolution and multi-view photometric stereo, but also aim at estimating camera poses, illumination, reflectance and high resolution geometry jointly, thus circumventing the need of tedious offline calibration of a freely moving camera or lighting.
As we will see later, our setup can be used out-of-the-box, i.e. we attach an LED light source to an RGB-D camera to enforce different illumination across different views while moving the camera.
\section{Variational Depth Super-Resolution and Multi-View Photometric Stereo}\label{sec3}
Our proposed model consists of three major parts and we will discuss each in more mathematical detail before putting all together to an end-to-end variational formulation. 
We start with some theoretical foundation on depth super-resolution, before formalizing an image formation model for photometric stereo.
An extension to multi-view PS is introduced using rigid body motions and how to naturally incorporate them to the classical PS model.

\subsection{Depth Super-Resolution}\label{sec:depth_sr}
Depth super-resolution tries to infer a high resolution (HR) depth map $\z:\OHR\subset\R^2\to\R$, from $n\geq1$ low-resolution (LR) depth measurements $\zi:\OLR\subset\R^2\to\R$, $i\in\mathcal{I}:=\{0,\dots,n-1\}$.
Formally, the corresponding forward process can be written as
\begin{equation}\label{eq:sr_forward}
\zi = D_i\z + \epsiloniz,\qquad\forall i\in\I,
\end{equation}
where $D_i:\OHR\to\OLR$ is some linear downsampling operator~\cite{unger2010} and $\epsiloniz(\p)\sim\mathcal{N}(0,\sigma_z^2)$, $\p\in\OLR$, the realization of a stochastic process following a Gaussian distribution with mean $0$ and variance $\sigma_z^2$, see~\cite{khoshelham2012} for more details.
If the cameras motion is not static, then these $n$ LR depth measurements are captured from different viewpoints and usually one needs to define a so called reference frame. 
Furthermore, notion of the camera movement needs to be taken into account in the process described in \eqref{eq:sr_forward}.
This can be realized in the linear operator $D_i$, such that not solely downsampling, but also image warping is considered~\cite{unger2010}.
Although we aim to tackle the problem depth super-resolution from multiple viewpoints, we will see later that our final variational model only depends on the LR reference depth, i.e. for a LR depth map we rely on a single observation (``$n=1$'') and denote it with $\zz$ (we also drop the dependency on $i$ in $D_i$).
The multi-view aspect in our model arises from the photometric constraint which we discuss in Section~\ref{sec:mv}.
To invert~\eqref{eq:sr_forward} we follow a bayesian rationale for our variational approach~\cite{mumford1994}, i.e. 
\eqref{eq:sr_forward} can be formulated as a minimization problem of the form
\begin{equation}\label{eq:sr_opt}
\min_{z:\OHR\to\R} \tau\Norm{Dz - \zz}^2_{\ell_2(\OLR)} + \Reg(z),
\end{equation}
where $\Norm{\cdot}^2_{\ell_2(\OLR)}$ is the euclidean $2$-norm over the image domain $\OLR$ and $\tau$ is a trade-off parameter determining the impact of each of the two summands in~\eqref{eq:sr_opt}.
The so called regularization term $\Reg(z)$ is often based on prior knowledge.
One choice is a total variation based regularization term incorporating sparse features of the intensity image~\cite{ferstl2013}. 
Recently, a promising direction towards depth refinement is to use not only sparse, but dense information from the RGB image, i.e. use photometric cues to perform depth refinement~\cite{haefner2019tpami,or2015}.
\subsection{Photometric Stereo}\label{sec:PS}
Let us consider the following image formation model under Lambertian reflectance to establish a dense relation between shape and intensity:
\begin{equation}\label{eq:ifm}
I_c(\p) = \rho_c(\p)\int_{\S^2}\phi(\bomega)\max(0,\dotp{\bomega,\n(\p)}\dx\bomega,
\end{equation}
where $I_c(\p),\rho_c(\p)\in\R$ is the image intensity, albedo in channel $c\in C=\{R,G,B\}$ at pixel $\p\in\OHR$, respectively.
The integral is referred to as shading and is carried out over the $3$-dimensional unit sphere $\S^2$, i.e. all possible lighting directions $\bomega$ with their corresponding intensity $\phi(\bomega)$.
$\n(\p)$ is the surface normal represented at pixel $\p\in\OHR$ and $\dotp{\cdot,\cdot}:\R^3\times\R^3\to\R$ is the $3$-dimensional dot product.
Using the well-known first-order spherical harmonic approximation which accounts for $87\%$ of real-world lighting~\cite{basri2003}, we can write \eqref{eq:ifm} as
\begin{equation}
I_c(\p) = \rho_c(\p)\dotp{\ml,\m(\n(\p))},
\end{equation}
where the shading term is simplified to a $4$-dimensional dot product with $\ml\in\R^4$ representing the global lighting of the scene and $\m(\n(\p))\in\R^4$ are the spherical harmonic basis functions at pixel $\p\in\OHR$, 
\begin{equation}
\m(\n(\p)) := \left[1, \n(\p)^\top\right]^\top.
\end{equation}
As our approach is running on RGB images, we stack the values for each channel to have
\begin{equation}
\mI(\p) = \mrho(\p)\dotp{\ml,\m(\n(\p))},
\end{equation}
with $\mI(\p) = [I_R(\p),I_G(\p),I_B(\p)]^\top$ being the RGB intensity values and $\mrho(\p) = [\rho_R(\p),\rho_G(\p),\rho_B(\p)]^\top$ being the RGB albedo values.
We have now established a mathematical relationship between the color image $\mI$ and the geometry of the scene described with the surface normals $\n$.
As we use (perspective) depth maps from an RGB-D sensor and we need to link absolute depth (, i.e. depth maps) with relative geometric information (, i.e. surface normals), we parametrize the normal-field $\n$ in terms of depth $z$ under perspective projection and assuming differentiability~\cite{queau2018jmiv},
\begin{equation}\label{eq:normal}
\n(\z(\p)) = \frac{1}{\dA(z(\p))}\begin{bmatrix}f\nabla z(\p)\\ -1-\dotp{\p-\pc,\nabla z(\p)}\end{bmatrix}\in\S^2,
\end{equation}
where $\dA(z(\p)) = \sqrt{\norm{f\nabla z(\p)}^2 + (1+\dotp{\p-\pc,\nabla z(\p)})^2}$ is the surface area element under perspective projection~\cite{graber2015}, $\nabla$ is the gradient operator and $\norm{\cdot}$ is the 2-dimensional euclidean norm, $f$ and $\pc=\left[c_x,c_y\right]^\top$ are camera parameters, i.e. the focal length and principal point, respectively.
To avoid clutter in notation we drop the argument dependency on the pixel $\p$ and rather consider each quantity on the whole image domain $\OHR$,
\begin{equation}
\mI = \mrho\dotp{\ml,\m[\n[z]]},
\end{equation}
with the RGB intensity image $\mI:\OHR\to\R^3$, the albedo $\mrho:\OHR\to\R^3$, the global lighting vector $\ml\in\R^4$, and the stacked spherical harmonic basis functions $\m$  depending on the surface normals $\n$ parametrized with $z$, i.e. we actually consider a vector-field $\m[\n[z]]:\OHR\to\R^4$.
In order to perform photometric stereo, $n\geq3$ images under different illumination need to be acquired~\cite{woodham1980}, and since geometry and reflectance are properties of the scene and stay fixed, these changes become only apparent in lighting and intensity, this can be mathematically realized with
\begin{equation}\label{eq:ps_forward}
\mIi = \mrho\dotp{\mli,\m[\n[z]]},\quad i\in\I.
\end{equation}
Practically, changes in illumination can be achieved by using a light stage, illuminating the scene with a light-bulb or LED light source, or capturing pictures during different times of the day or even during different seasons of the year~\cite{abrams2012,ackermann2012,jung2019}.
\\
Based on \eqref{eq:sr_opt} and \eqref{eq:ps_forward}, it was shown in~\cite{haefner2019tpami,peng2017} that under the assumption of $\{\mIi\}_{i\in\I}$ being corrupted with homoskedastically Gaussian-distributed noise a purely data-driven, photometric regularization term can be established,
$\Reg_{\cite{peng2017}}(z)=\sum_{i\in\I}\Norm{\mIi - \mrho\dotp{\mli,\m[\n[z]]}}^2_{\ell_2(\OHR)}$.
Apart from avoiding a man-made regularization term, both, the depth prior term (left summand in \eqref{eq:sr_opt}) and photometric term $\Reg_{\cite{peng2017}}(z)$ can be each others helping hand.
Treated separately, depth super-resolution and uncalibrated photometric stereo are ill-posed problems by itself~\cite{basri2007,belhumeur1999,farsiu2004}, but the combination of both can help disambiguating the other:
The depth prior term resulting from an RGB-D sensors initial depth uniquely defines the Lorentz transformation in order to disambiguate uncalibrated photometric stereo~\cite{brahimi2019}.
Similarly, 
using the 
information comprised in the intensity image and thus in the photometric term can help to result in fine detailed depth estimates and to disambiguate depth super-resolution.
To this end it seems intuitive to combine both approaches to get the best out of both worlds.
Yet,~\cite{peng2017} assumes a static camera-to-object setup during the capturing process which makes it impractical to use in a scenario where camera motion is present. 
A natural way to capture data would be to attach the light source to the RGB-D sensor and freely move the camera around.
This results in an easy way to capture data from different viewpoints while simultaneously changing illumination. 
Furthermore, resorting to RGB-D cameras has the advantage of robustly capturing depth data, independently of illumination changes in the visible spectrum, which gives coarse, but robust knowledge of the geometry of the scene which can be used during inference.
The next section is concerned with how to incorporate the practical setup of attaching an LED light source to an RGB-D camera in the variational formulation.
To this end, rigid body motions and how we will use them to model multi-view photometric stereo are discussed.

\subsection{Using Rigid Body Motion for Multi-View Photometric Stereo}\label{sec:mv}
Rigid body motions can be used to describe how a camera moves in space, and additionally gives a relation between pixel positions of different images captured from different viewpoints.
Under perspective projection a pixel $\p\in\OHR\subset\R^2$ and its depth value $z(\p)$ is used to express its conjugate 3D point $\P\in\R^3$ in the scene,
\begin{equation}\label{eq:reprojection}
\pi^{-1}(\p,\z(\p)) = z(\p)\begin{bmatrix}f^{-1}(\p-\pc)\\1\end{bmatrix} = \P,
\end{equation}
where $\pi^{-1}:\OHR\times\R\to\R^3$ can be referred to as the reprojection function.
The relation between $\P$ seen from one camera and the same corresponding point $\tilde\P$ seen from a second camera can be expressed as a rigid body motion.
This comprises a rotation $\Rot\in SO(3)$, where $SO(3)$ is the space of $3\times3$ dimensional rotation matrices, and a translation $\trans\in\R^3$,
\begin{equation}\label{eq:coord_trans}
\T[\Rot,\trans](\P) = \Rot\P + \trans = \tilde\P,
\end{equation}
with $\T[\Rot,\trans]:\R^3\to\R^3$.
After applying the coordinate transformation with $\T[\Rot,\trans]$, the resulting 3D point $\tilde\P=(\tilde X,\tilde Y,\tilde Z)^\top$ has to be projected on the image plane again,
\begin{equation}\label{eq:projection}
\pi(\tilde\P) = \frac{f}{\tilde Z} \begin{bmatrix}\tilde X\\\tilde Y\end{bmatrix} - \pc = \tilde\p,
\end{equation}
where $\pi:\R^3\to\OHR$.
We can describe the process of re-projection \eqref{eq:reprojection}, camera coordinate transformation \eqref{eq:coord_trans} and projection \eqref{eq:projection} using a single warping function $\bw[\Rot,\trans,\z]:\OHR\to\OHR$,
\begin{equation}\label{eq:warping}
\bw\left[\Rot,\trans,\z\right]\left(\p\right) = \pi\left( \T\left[\Rot,\trans\right]\left(\pi^{-1}\left(\p,z\left(\p\right)\right) \right) \right).
\end{equation}
This warping function describes how a pixel from one image frame is transformed to locate the same 3D point in the second image.
As we have multiple (namely $n\geq3$) observations due to the multi-view setup, we have to keep track of each rotation and translation.
To this end, we transform the camera coordinate system from the reference frame ($0$-th frame, cf. Section~\ref{sec:depth_sr}) to the $i$-th frame using the rotation $\Roti$ and translation $\transi$.
Note that this transformation needs only the depth map from the reference frame, as the re-projection is applied to the reference frame, thus only the rigid body motion is depending on the camera pose of the $i$-th frame.
Photometrically, under the Lambertian assumption the intensity in the $i$-th frame at pixel $\tilde\p_i$ can be described as illuminating the scene from the $i$-th frame, but seen from the reference frame at pixel $\p$, mathematically this can be written by combining \eqref{eq:ps_forward} and \eqref{eq:warping} as
\begin{equation}\label{eq:mv_ps_forward}
\mIi\circ \bw[\Roti,\transi,\z] = \mrho\dotp{\mli,\m[\n[z]]},\quad i\in\I.
\end{equation}
For simplicity, we define $\tilde\mIi[\Roti,\transi,\z]:=\mIi\circ \bw[\Roti,\transi,\z]$ and use this representation throughout the rest of this work, before discussing our proposed variational setup.

\subsection{Proposed Variational Formulation}
Putting together the forward model on depth super-resolution~\eqref{eq:sr_forward} and multi-view uncalibrated photometric stereo~\eqref{eq:mv_ps_forward}, we formulate the joint recovery of camera motion, geometry, reflectance and lighting as the following variational problem
\begin{align}\label{eq:proposed_model}
\min_{\substack{\{\Roti\}\in SO(3)\\ \{\transi\}\in\R^3\\\z:\OHR\to\R\\\mrho:\OHR\to\R^3\\\{\mli\}\in\R^4}}&
\sum_{i\in\I}\Phi_\C\left( \tilde\mIi\left[\Roti,\transi,\z\right] - \mrho\dotp{\m\left[\n\left[z\right]\right],\mli} \right)\nonumber\\[-3em]
& + \tau\Norm{D\z - \zz}^2_{\ell_2(\OLR)}.\\[-0.5em]\nonumber
\end{align}
We denote with $\Phi_\C(\cdot)$ the sum over all pixels and all channels over data-fitting discrepancy,
\begin{multline}\label{eq:mestimator}
\Phi_\C\left(\tilde\mIi\left[\Roti,\transi,\z\right] - \mrho\dotp{\m\left[\n\left[z\right]\right],\mli}\right) =\\
\sum_{\p\in\OHR}\!\sum_{c\in C}\! \phi_\C\!\left(\!\mtIic\!\left[\Roti,\!\transi,\!\z\right]\!\left(\p\right)\! -\! \mrhoc\!\left(\p\right)\!\dotp{\m\!\left[\n\!\left[z\right]\right]\!\left(\p\right)\!,\!\mli}\right),
\end{multline}
where $\phi_\C(r)=\frac{\lambda^2}{2}\log\left(1+\frac{r^2}{\lambda^2}\right)$ is Cauchy's M-estimator (evaluated at residual $r$) to make the approach robust against outliers.
In the context of photometric stereo outliers can be self- or cast-shadows, specularities, and image noise or quantization, see~\cite{queau2017cvpr} for more detail.
The parameter $\lambda$ is a tunable hyper-parameter and we heuristically found $\lambda=0.04$ to be a good choice.
We will now discuss our numerical approach to solve the variational problem on depth super-resolution using multi-view uncalibrated PS.
\section{Numerical Resolution}\label{sec4}
The problem depicted in~\eqref{eq:proposed_model} is nonconvex wrt. $z$ (due to the normalization factor, cf.~\eqref{eq:normal}), ${\Roti}$, as well as Cauchy's M-estimator.
This makes our variational problem difficult to solve and we will now discuss a strategy to make optimization tractable.
To this end, we follow~\cite{haefner2019iccv,peng2017} and rewrite the normals in terms of its \textit{unnormalized normal} $\tilde\n\left[z\right]$ and its norm, $\n\left[z\right] = \frac{\tilde\n\left[z\right]}{\norm{\tilde\n\left[z\right]}} = \frac{\tilde\n\left[z\right]}{\dA\left[z\right]}$, this allows us to split the nonconvexity induced by the normal into its convex (linear) and its nonconvex part.
Camera rotation and translation represent elements of the Lie group $SE(3)$, which for the numerical solution we parameterize by the corresponding Lie algebra element, i.e. the $6$-dimensional twist coordinates $\bxii\in\R^6$, $\Roti=\Roti\left(\bxii\right)$ and $\transi=\transi\left(\bxii\right)$. This allows to optimize over the Euclidean parameters $\bxii$, cf.~\cite{kerl2013,ma2012}.
We can now rewrite~\eqref{eq:proposed_model} as
\begin{align}\label{eq:proposed_numerical_model}
\min_{\substack{\{\bxii\}\in\R^6\\\z:\OHR\to\R\\\mrho:\OHR\to\R^3\\\{\mli\}\in\R^4}}&
\!\sum_{i\in\I}\!\Phi_\C\!\!\left(\! \tilde\mIi\!\left[\Roti\!\left(\bxii\right)\!,\!\transi\!\left(\bxii\right)\!,\!\z\right]\! -\! \mrho\!\dotp{\!\m\!\left[\frac{\tilde\n\left[z\right]}{\dA\left[z\right]}\!\right]\!,\!\mli\!}\! \right)\nonumber\\[-2.5em]
& + \tau\Norm{D\z - \zz}^2_{\ell_2(\OLR)}.\\[-0.5em]\nonumber
\end{align}
The nonconvexity induced by $\phi_\C$ in~\eqref{eq:mestimator} gives rise to perform a reweighted least squares optimization scheme, cf.~\cite{queau2017cvpr}.
This boils down to approximating $\phi_\C$ at the $k$-th iteration and residual $r^{(k)}$ with its weighted least squares estimator, $\phi_\C\left(r^{(k)}\right)\approx\phi_\C^{(k)}\left(r^{(k)}\right):=w^{(k)}\left(r^{(k)}\right)\phi_2\left(r^{(k)}\right)$, where $w^{(k)}\left(r^{(k)}\right)=\frac{\phi'_C\left(r^{(k)}\right)}{r^{(k)}}$ and $\phi_2\left(r^{(k)}\right)=\norm{r^{(k)}}^2$.
Analogously, we denote this approximation at the $k$-th iteration 
with $\Phi_\C^{(k)}$.
That is, at the $(k)$-th iteration with the given estimates $(\{\bxii^{(k)}\}, \z^{(k)}, \mrho^{(k)}, \{\mli^{(k)}\})$ we update each quantity according to the following sweep:
\begin{align}\label{eq:xi_update}\newcommand{\myspace}{-0.em}
&\bm{\xi}_i^{(k+1)} \!=\argmin_{\bm{\xi}_i\in\R^6} &&\hspace*{-0.75em}\! \Phi_\C^{(k)}\!\bigg(\!\tilde\mIi\!\left[\Roti\!\left(\bxii\right)\!,\!\transi\!\left(\bxii\right)\!,\!\z^{(k)}\!\right]
\\[-0em]
&&&\hspace*{-0.75em}\! -\! \mrho^{(k)}\!\dotp{\!\m\!\left[\!\frac{\tilde\n\left[z^{(k)}\right]}{\dA\left[z^{(k)}\right]}\!\right]\!,\!\mli^{(k)}\!}\! \bigg),\,\forall i\in\I,\nonumber
\end{align}\vspace*{-1.1em}
\begin{align}\label{eq:depth_update}
&z^{(k+1)} \!= \argmin_{z:\OHR\to\R}&&\hspace*{-0.75em}\!\sum_{i\in\I}\!\Phi_\C^{\left(k+\frac{1}{4}\right)}\!\bigg( \!\tilde\mIi\!\left[\Roti\!\left(\!\bxii^{(k+1)}\!\right)\!,\!\transi\!\left(\!\bxii^{(k+1)}\!\right)\!,\!\z^{(k)}\!\right]\nonumber
\\[-0em]
&&&\hspace*{-0.75em}\! -\! \mrho^{(k)}\!\dotp{\!\m\!\left[\!\frac{\tilde\n[z]}{\dA\left[z^{(k)}\right]}\!\right]\!,\!\mli^{(k)}\!}\! \bigg)
\\[-0em]
&&&\hspace*{-0.75em} + \tau\Norm{D\z - \zz}^2_{\ell_2(\OLR)},\nonumber
\end{align}\vspace*{-1.9em}
\begin{align}\label{eq:albedo_update}
&\mrho^{(k+1)} \!=\!\!\argmin_{\mrho:\OHR\to\R^3}&&\hspace*{-1.25em}\sum_{i\in\I}\!\Phi_\C^{\left(k+\frac{2}{4}\right)}\!\bigg(\! \tilde\mIi\!\left[\!\Roti\!\left(\!\bxii^{(k+1)}\!\right)\!,\!\transi\!\left(\!\bxii^{(k+1)}\!\right)\!,\!\z^{(k+1)}\!\right]\nonumber
\\[-0em]
&&&\hspace*{-1.75em} \!-\! \mrho\!\dotp{\!\m\!\left[\!\frac{\tilde\n\left[z^{(k+1)}\right]}{\dA\left[z^{(k+1)}\right]}\!\right]\!,\!\mli^{(k)}}\! \bigg),
\end{align}\vspace*{-1.1em}
\begin{align}\label{eq:lighting_update}
& \mli^{(k+1)} \!= \argmin_{\mli\in\R^4}&&\hspace*{-0.75em}\!\Phi_\C^{\left(k+\frac{3}{4}\right)}\!\bigg(\! \tilde\mIi\!\left[\!\Roti\!\left(\!\bxii^{(k+1)}\!\right)\!,\!\transi\!\left(\!\bxii^{(k+1)}\!\right)\!,\!\z^{(k+1)}\!\right]\nonumber
\\[-0em]
&&&\hspace*{-0.75em} \!-\! \mrho^{(k+1)}\!\dotp{\!\m\!\left[\!\frac{\tilde\n\left[z^{(k+1)}\right]}{\dA\left[z^{(k+1)}\right]}\!\right]\!,\!\mli}\! \bigg),\,\forall i\in\I.
\end{align}
We follow a coarse-to-fine strategy with five pyramid levels, i.e. on the coarsest level the resolution is $\frac{1}{16}$ of the original size.
For each pyramid level we solve~\eqref{eq:proposed_numerical_model} using~\eqref{eq:xi_update} -- \eqref{eq:lighting_update}.
The estimated quantities are then (bilinearly upsampled for $z$ and $\mrho$) used as input for the next pyramid level.
We initialize $\bxi_0^{(0)}\equiv\bm{0}$ (identity), and $\bxi_{i+1}^{(0)} = \bxi_{i}^{(1)}$, $i\in\I$ (small baseline assumption).
$z^{(0)}$ and $\mrho^{(0)}$ are bilinearly downsampled versions of $\zz$ and $\mI_0$, respectively and $\mli^{(0)}=(0.2,0,0,-1)$ (little ambient amount and frontal lighting), $\forall i\in\I$. 
Our algorithm has converged if the relative energy falls below a value of $10^{-5}$, which on average took 20 iterations for each pyramid level.
Note that \eqref{eq:depth_update} -- \eqref{eq:lighting_update} are linear least square problems, while we resort to Gauss-Newton iterations for~\eqref{eq:xi_update}, cf.~\cite{kerl2013}.
Next, we quantitatively and qualitatively evaluate our approach 
on synthetic and real-world data.

\section{Empirical Validation}\label{sec5}
We validate our method on synthetic and real-world data.
We draft insightful experiments of the involved parameters and compare against other state-of-the-art depth SR approaches.

\subsection{Synthetic Data}
\begin{figure*}[!ht]
  \centering
  \setlength\tabcolsep{0pt} 
  \def\arraystretch{0.2125} 
  \newcommand{\mywidthx}{0.125\textwidth} 
  \newcommand{\mywidthy}{0.25\textwidth} 
  \newcommand{\mywidthlr}{0.08\textwidth} 

  \newcolumntype{X}{>{\centering\arraybackslash}m{\mywidthx}}
  \newcolumntype{Y}{>{\centering\arraybackslash}m{\mywidthy}}
  \begin{tabular}{YYXXXX}
    ``Joyful Yell'' & ``Stanford Bunny'' & \multicolumn{2}{c}{``Joyful Yell''}& \multicolumn{2}{c}{``Stanford Bunny''}\\
    \multirow{3}{*}[2.25em]{\includegraphics[width=0.25\textwidth]{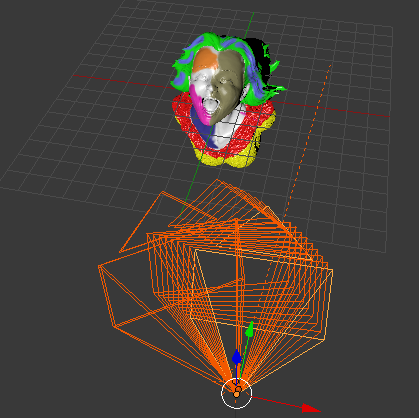}}&
    \multirow{3}{*}[2.25em]{\includegraphics[width=0.25\textwidth]{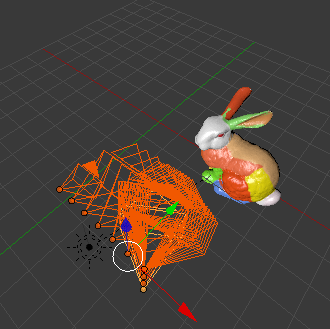}}&
    \includegraphics[height=\mywidthx]{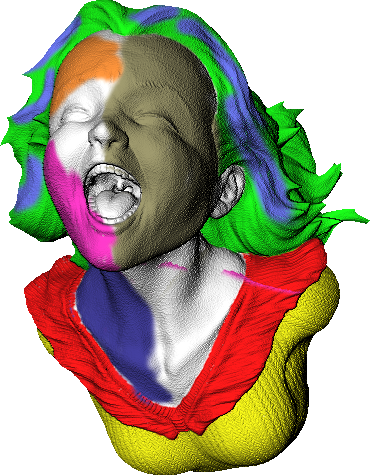}&
    \includegraphics[height=\mywidthx]{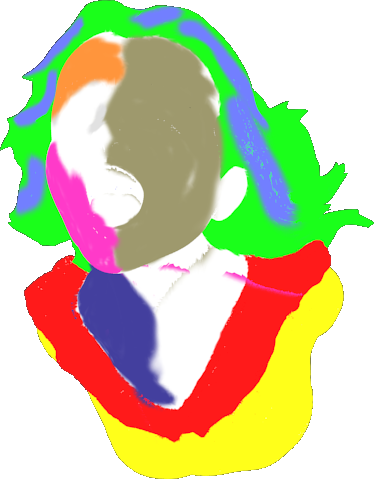}&
    \includegraphics[height=\mywidthx]{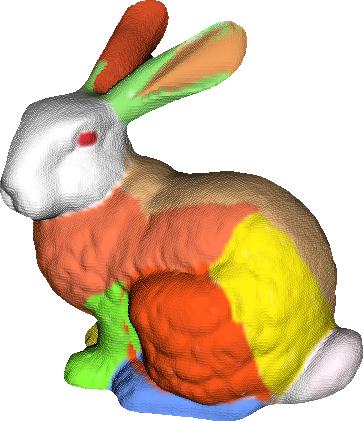}&
    \includegraphics[height=\mywidthx]{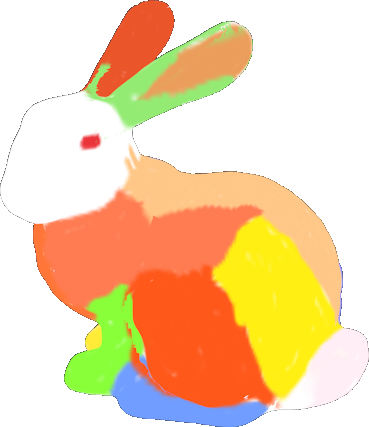}\\
     & &$\mI_0$ & $\mrho$ & $\mI_0$ & $\mrho$\\
    &
    &
    \includegraphics[height=\mywidthlr]{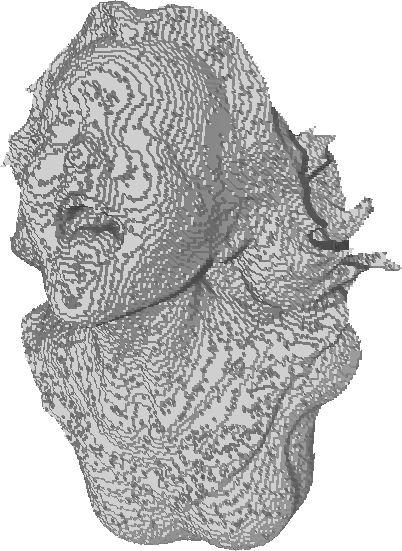}&
    \includegraphics[height=\mywidthx]{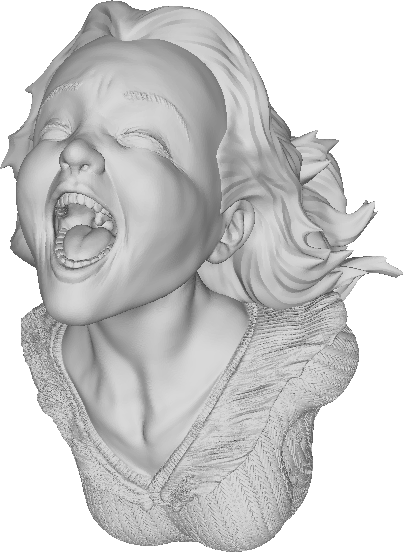}&
    \includegraphics[height=\mywidthlr]{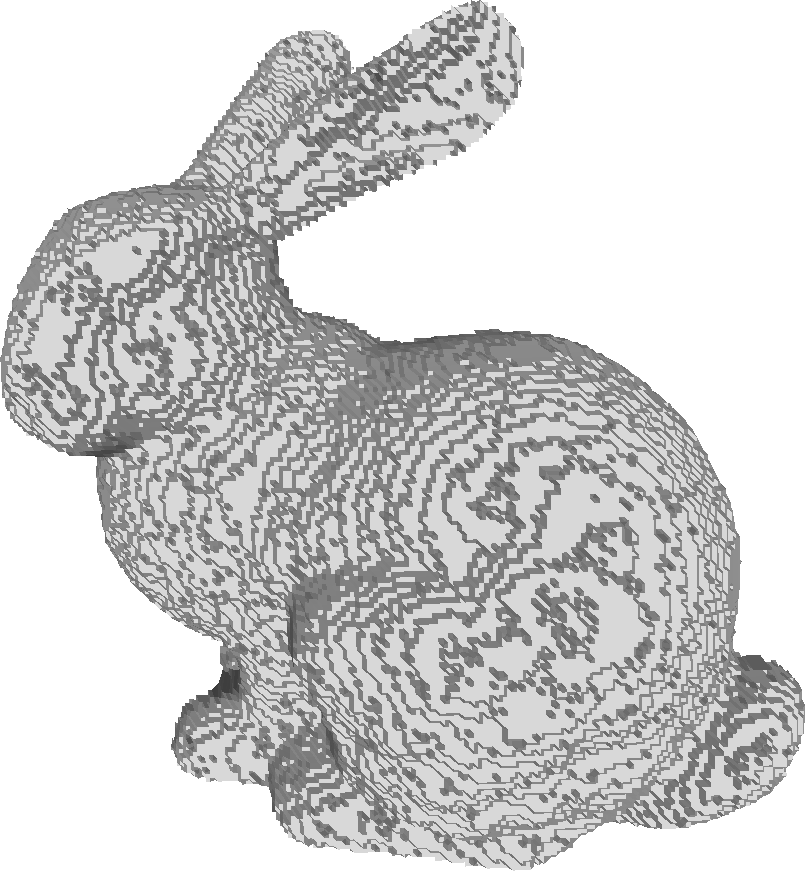}&
    \includegraphics[height=\mywidthx]{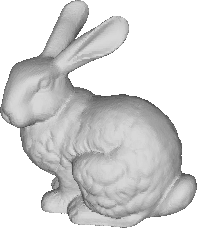}\\
    \multicolumn{2}{c}{3D view visualizing camera motion for both datasets.} & $\zz$ & $\z$ & $\zz$ & $\z$
  \end{tabular}
  
  \caption{(Left) Visualization of ``Joyful Yell'' and ``Stanford Bunny'' in space with corresponding camera motion used to render both image sequences.
  The motion for the ``Joyful Yell'' dataset comprises rotation only, the ``Stanford Bunny'' got captured under rotation and translation.
  The experiments will show that our approach handles both camera motions with similar ease and precision (quite in contrast to passive sensing approaches where depth cannot be recovered under pure rotation).
  (Right) The rendered reference image, ground-truth albedo, low-resolution reference depth and ground-truth depth.
  }
  \label{fig:synthetic_data}
\end{figure*}

We use two publicly available datasets for synthetic evaluation, ``Joyful Yell''~\cite{bendansie2015} and `` Stanford Bunny''~\cite{turk1994}.
Both objects are used and colored within the rendering software Blender.
A virtual camera with collocated light source is used to render images from different viewpoints.
The resulting data, with its corresponding ground truth is then used for evaluation.
Low-resolution depth maps ($\frac{1}{2},\frac{1}{4},\frac{1}{8}$ of original size) are generated by adding non-uniform zero-mean Gaussian noise with standard deviation $10^{-5}$ times the squared original depth value~\cite{khoshelham2012}, cf. Figure~\ref{fig:synthetic_data} for an illustration.

\begin{figure}[!ht]
  \centering
  \setlength\tabcolsep{1pt} 
  \def\arraystretch{0} 
  \newcommand{\mywidth}{0.235\textwidth} 

  \newcolumntype{X}{>{\centering\arraybackslash}m{\mywidth}}
  \begin{tabular}{XX}
    \includegraphics[width=\mywidth]{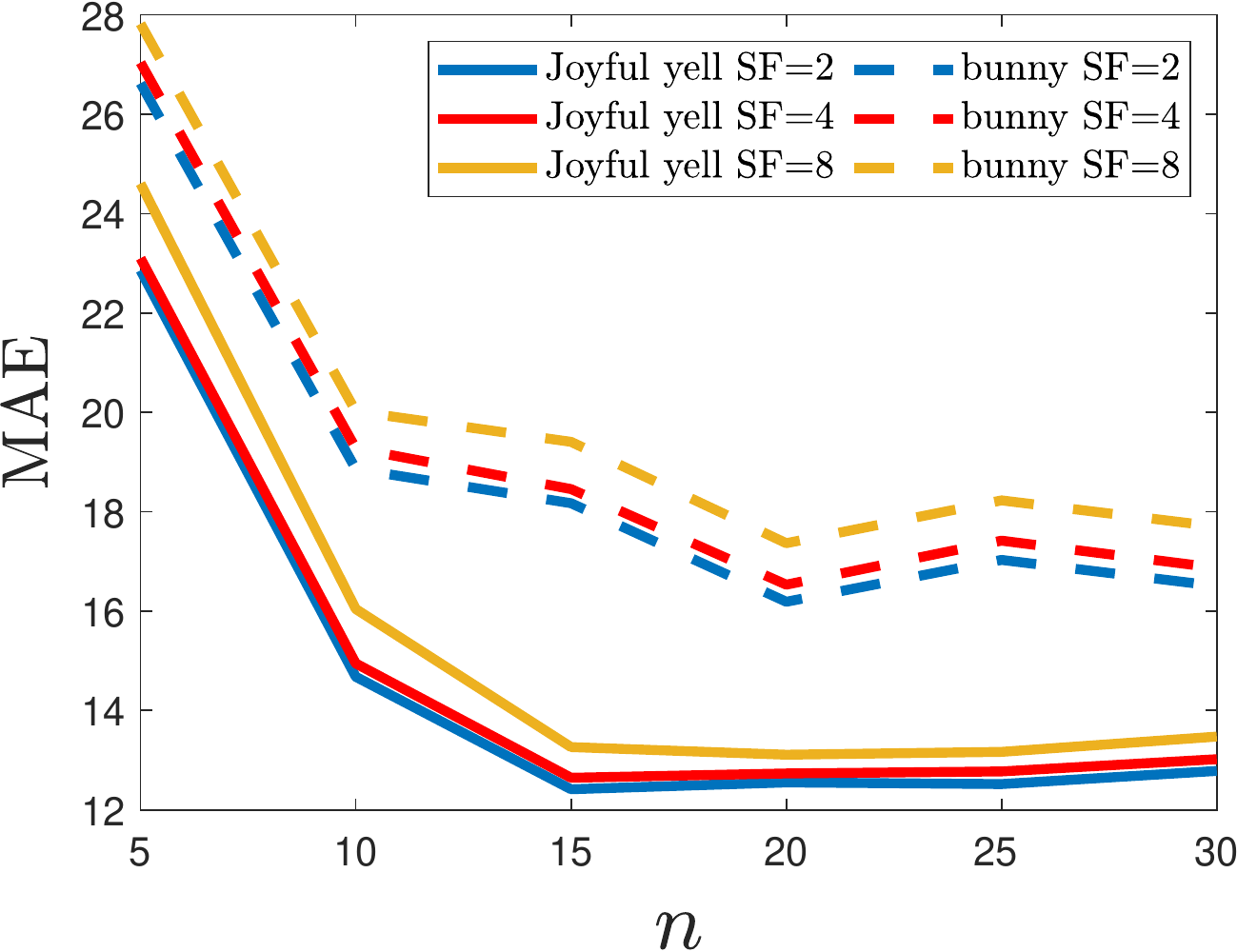}&
    \includegraphics[width=\mywidth]{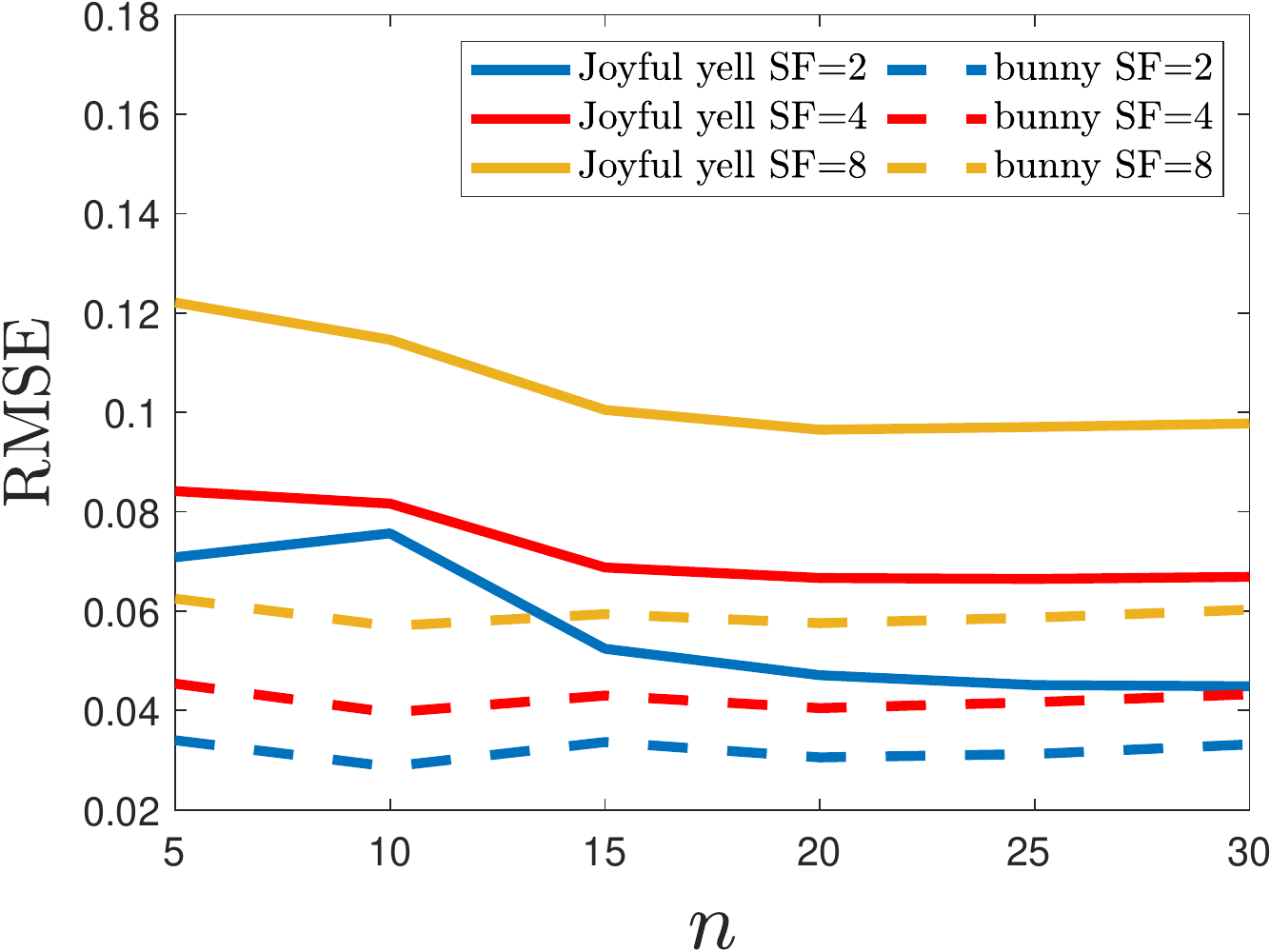}
  \end{tabular}
  \caption{Impact of the number of input images $n$ on the accuracy for three scaling factors (SF).
  Left is the Mean Angular Error (MAE, in degrees) on the normals.
  Right is the Root Mean Square Error (RMSE, in arbitrary units) on depth.
  $15$ -- $20$ images are enough to obtain accurate results.
  More than $20$ images seem to increase the error, due to a larger baseline to the reference frame.}
  \label{fig:num_images}
\end{figure}

While having ground-truth at hand using synthetic data, we evaluate how $n$ and $\tau$ impacts the mean angular error (MAE) and the root mean squared error (RMSE).
\\
\textbf{Parameter Evaluation.}
To evaluate for $n$, we use the first $n\in\{5,10,15,20,25,30\}$ frames of the image sequence.
The results obtained with our proposed approach are reported in Figure~\ref{fig:num_images}.
With increasing number of images the error in both error metrics is decreasing.
Yet, for more than $15$--$20$ images the error increases again.
This is most likely due to a larger baseline between the images and the $0$-th frame, error is accumulated in camera pose estimation and thus depth inference suffers after warping.
Given this insight, we choose $n=20$ for all our experiments.

\begin{figure}[!ht]
  \centering
  \setlength\tabcolsep{1pt} 
  \def\arraystretch{0} 
  \newcommand{\mywidth}{0.235\textwidth} 

  \newcolumntype{X}{>{\centering\arraybackslash}m{\mywidth}}
  \begin{tabular}{XX}
    \includegraphics[width=\mywidth]{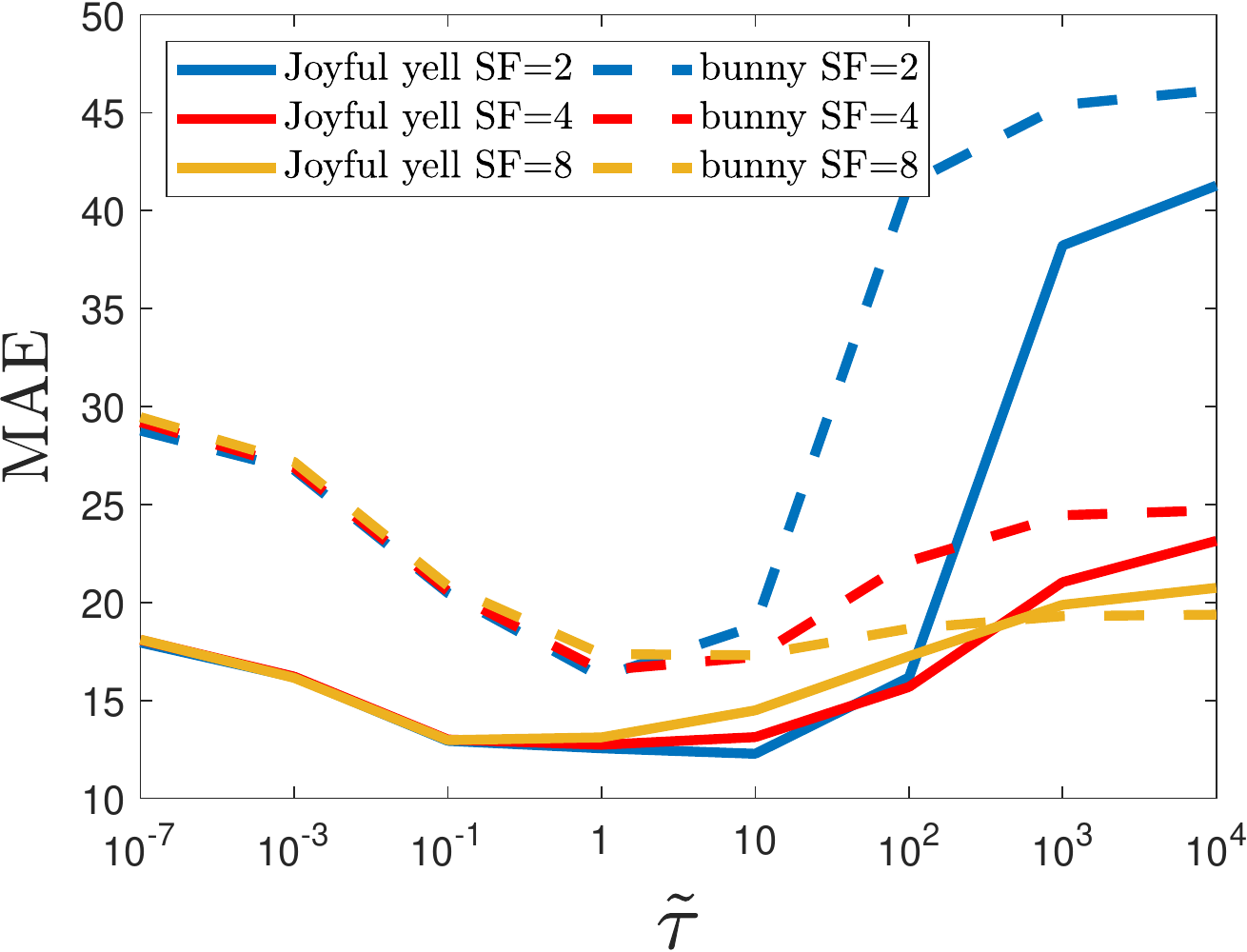}&
    \includegraphics[width=\mywidth]{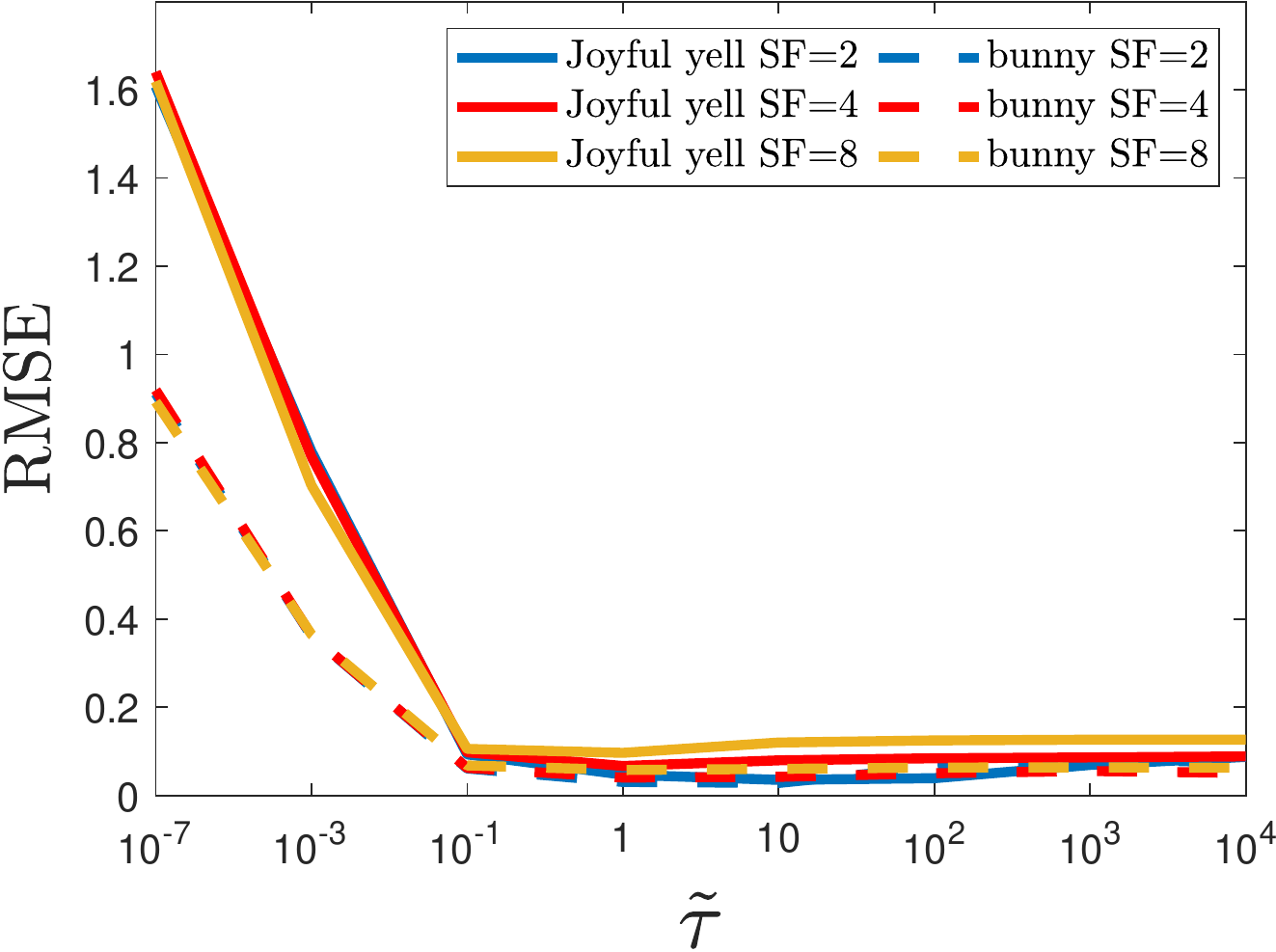}
  \end{tabular}
  \caption{Impact of the input parameter $\tilde\tau$ on the accuracy for three scaling factors (SF) under two different datasets.
Left is the Mean Angular Error (MAE, in degrees) on the normals.
Right is the Root Mean Square Error (RMSE, in arbitrary units) on depth.
The interval $\tau \in [1, 100] $ leads globally accurate results (low RMSE) with fine geometric details (low MAE).}
  \label{fig:parameter_tuning}
\end{figure}

Next, we evaluate against $\tau$.
Due to the pyramid scheme and to be more robust across datasets (larger/smaller objects, different depth magnitude, etc.) we perform parameter normalization, $\tau = \frac{n \mean(\{\mIi\})^2\norm{\OHR}\norm{C}}{\mean(\zz)^2\norm{\OLR}}\tilde\tau$, where $\mean(\cdot)$ is the mean operator, across all input RGB images or the low-resolution depth, $\norm{\cdot}$ here refers to the cardinality of $\OHR$, $C$ and $\OLR$ and $\tilde\tau$ is the parameter we evaluate against.
Results are shown in Figure~\ref{fig:parameter_tuning}.
For low values of $\tilde\tau$ the depth prior term can be considered to have little to no influence and the approach can be compared to an uncalibrated PS one.
Thus, the overall shape (RMSE is large) suffers from the ill-posedness of uncalibrated PS~\cite{basri2007}.
On the other hand, large values of $\tilde\tau$ result in a pure depth super-resolution approach, and the result can only get as good as the input depth, i.e. missing fine geometric information.
This is resembled in the MAE value, which heavily increases for too large values of $\tilde\tau$.
Given this insight, $\tilde\tau\in[1,100]$ seems a good trade-off between fine-scale details (low MAE) and global geometry fit (low RMSE), and we set $\tilde\tau=10$ in all further experiments.

Note that even for scaling-factor (SF) $8$ the difference in MAE and RMSE is small compared SF $4$ or $2$, showing that our approach is robust across different SF.

\paragraph{Comparison with other Methods.}
Having the parameters fixed, we quantitatively and qualitatively evaluate our approach against other state-of-the-art depth super-resolution approaches.
We compare to a mathematically principled robust optimization framework for color guided depth map restoration~\cite{liu2017}.
To show the impact of jointly solving for camera pose and shape, reflectance and lighting, we compare against~\cite{peng2017}, for which we first solve for camera poses (basically solving only~\eqref{eq:xi_update} with our framework by setting $z=\zz$ and $\mrho\dotp{m[n[z]],\mli}=\mI_0$) followed by applying the model of~\cite{peng2017} to the warped images.
In~\cite{li2016} a unified approach to a hierarchical optimization framework and fast global smoothing performing cascaded interpolation with alternating RGB guidance is used.
To address the impact of the multi-view setup, we also compare to a single-shot depth super-resolution scheme incorporating photometric cues using the shape-from-shading principle~\cite{haefner2018cvpr}.
Table~\ref{tab:synthetic_quantitative} shows quantitative (on Joyful Yell and Stanford Bunny) and Figure~\ref{fig:comparison} shows qualitative (on Joyful Yell, Stanford Bunny and real-world data we captured ourselves) results.
In numbers, our approach performs best in terms of both error metrics, showing the superiority compared to other state-of-the-art methods.
This can also be verified visually, while the results of \cite{li2016} are too smooth, the ones of~\cite{liu2017} appear too noisy.
Better shape estimates can be achieved with~\cite{haefner2018cvpr,peng2017}, showing the positive impact on photometric approaches for depth super-resolution.
\begin{table}[]
  \centering
  \setlength\tabcolsep{1pt} 
  \def\arraystretch{1} 
  \begin{tabular}{|c|c|c|c|c|c|c|}
  \hline
  \multicolumn{1}{|c|}{Dataset}   & Metric & \cite{liu2017} & \cite{peng2017} & \cite{li2016} & \cite{haefner2018cvpr} & Ours \\ \hline
  \multirow{2}{*}{Joyful Yell}    & MAE    & 41.43  & 12.73 &  26.32 &   27.42& \textbf{12.54}  \\ \cline{2-7} 
                                  & RMSE   &  0.062  & 0.062  &  0.450  &   0.099 &\textbf{0.039} \\ \hline
  \multirow{2}{*}{Stanford Bunny} & MAE    &  32.33  & 21.42 &   20.89 &   25.86 &\textbf{16.06}  \\ \cline{2-7} 
                                  & RMSE   &   0.028 &    0.028 &   0.114 &       0.044 &\textbf{0.022}  \\ \hline
  \end{tabular}
  \caption{Quantitative results on the Joyful Yell and Stanford Bunny of our proposed method against other state-of-the-art depth super-resolution approaches.}
  \label{tab:synthetic_quantitative}
\end{table}

Next, we discuss our hardware setup and conclude the experimental section with a comparison on real-world results.

\subsection{Setup and Real-World Data}
\begin{figure}[h]
\centering
\includegraphics[width=\linewidth]{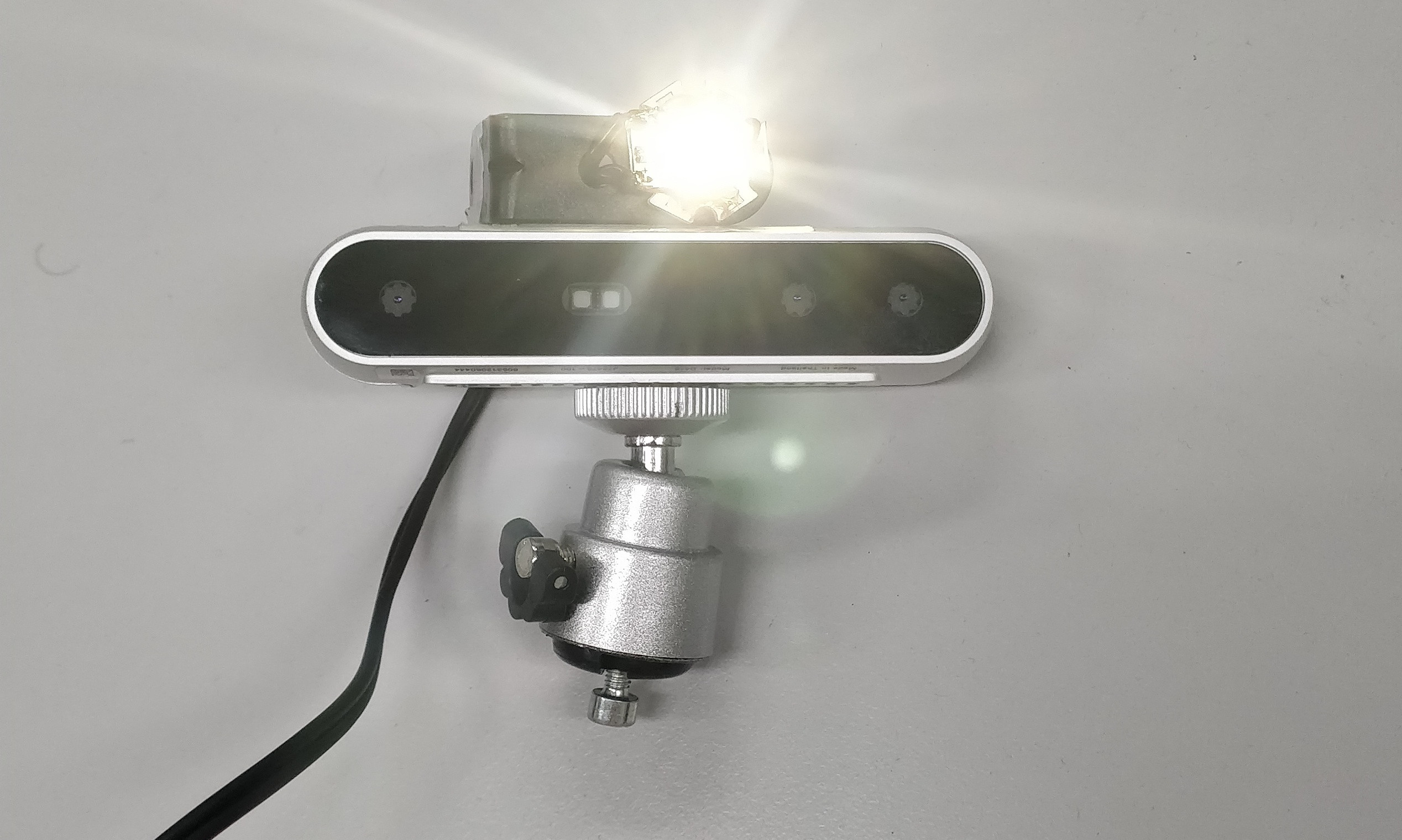}
\caption{Our setup we used to capture real-world data.
An LED light source is attached to an Intel RealSense D415 RGB-D camera.
We then move the camera to capture the objects of interest, see Figure~\ref{fig:comparison}.
Different illumination is induced by camera motion and the attached LED.}
\label{fig:setup}
\end{figure}

\begin{figure*}[!ht]
  \centering
  \setlength\tabcolsep{4pt} 
  \def\arraystretch{0} 
  \newcommand{\mywidth}{0.12\textwidth} 
  \newcommand{\mywidthlr}{0.08\textwidth} 
  \newcommand{\mywidths}{0.10\textwidth}

  \newcolumntype{C}{>{\centering\arraybackslash}m{0.06\textwidth} }
  \newcolumntype{X}{>{\centering\arraybackslash}m{\mywidth}}
  \newcolumntype{Y}{>{\centering\arraybackslash}m{\mywidthlr}}
  \begin{tabular}{CXYXXXXXX}
 & Reference Image $\mI_0$ & Low-resolution Depth $\zz$ & \cite{liu2017} & \cite{peng2017} & \cite{li2016} & \cite{haefner2018cvpr} & Ours \\
  \rotatebox{90}{Joyful Yell}&
  \includegraphics[width=\mywidth]{synthetic/joy1.png}&
  \includegraphics[width=\mywidthlr]{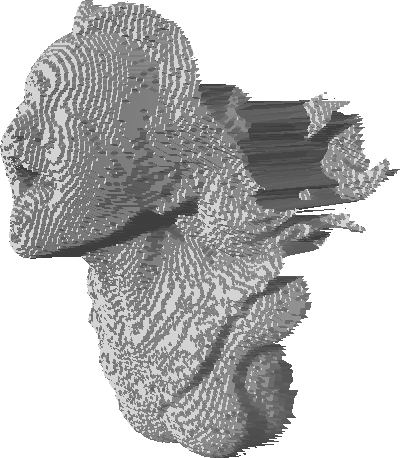}&
  \includegraphics[width=\mywidth]{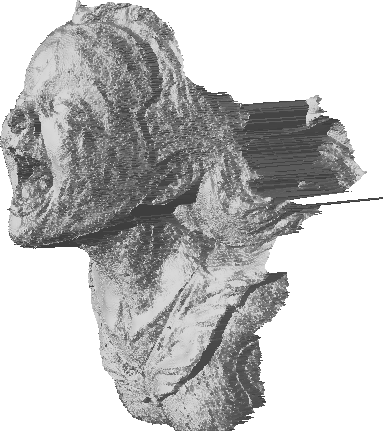}&
  \includegraphics[width=\mywidth]{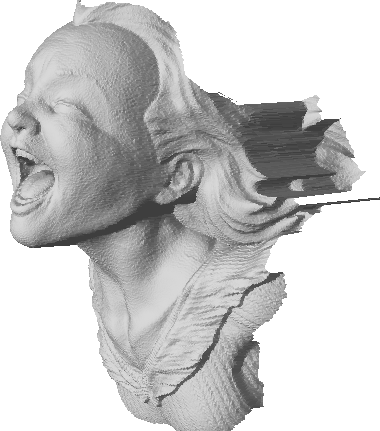}&
  \includegraphics[width=\mywidth]{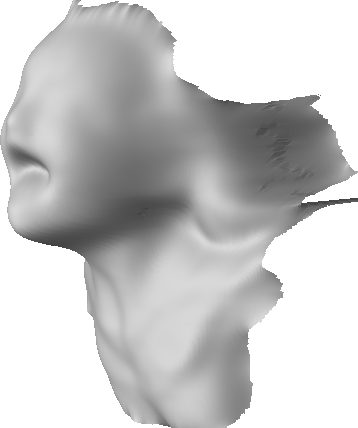}&
  \includegraphics[width=\mywidth]{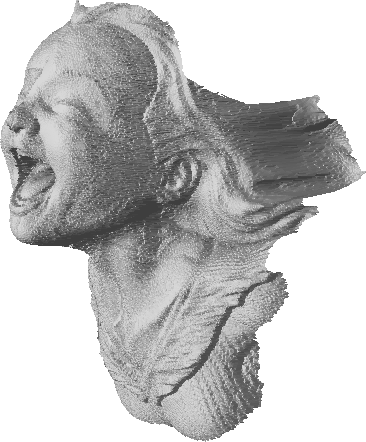} &
  \includegraphics[width=\mywidth]{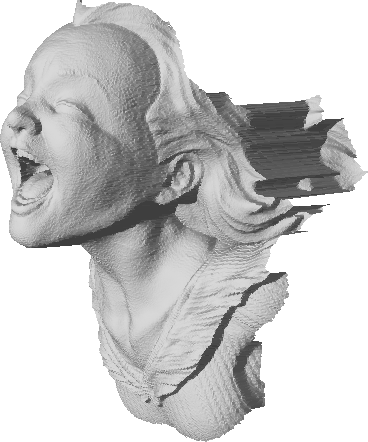}\\
  \rotatebox{90}{Stanford Bunny}&  
  \includegraphics[width=\mywidth]{synthetic/bunny_13.png}&
  \includegraphics[width=\mywidthlr]{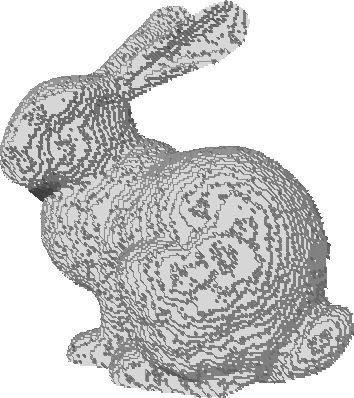}&
  \includegraphics[width=\mywidth]{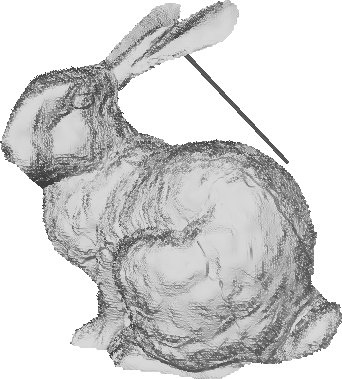}&
  \includegraphics[width=\mywidth]{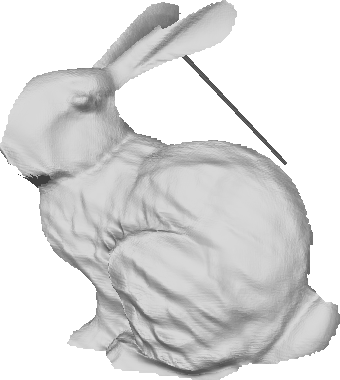}&
  \includegraphics[width=\mywidth]{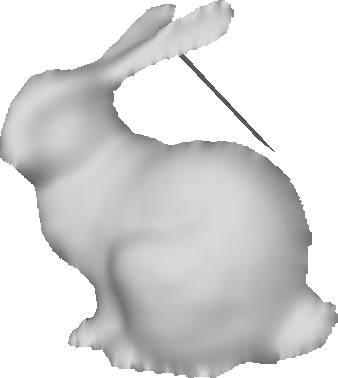}&
  \includegraphics[width=\mywidth]{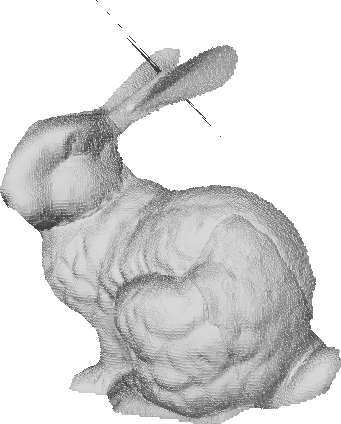}&
  \includegraphics[width=\mywidth]{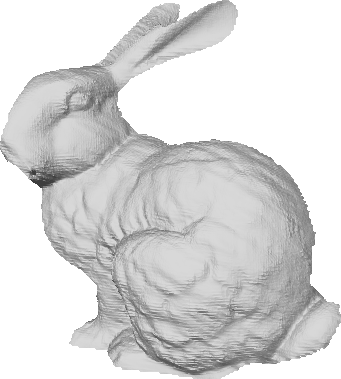}\\
   \rotatebox{90}{Jacket}&
  \includegraphics[width=\mywidth]{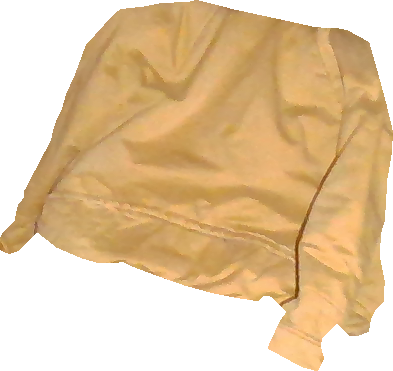}&
  \includegraphics[width=\mywidthlr]{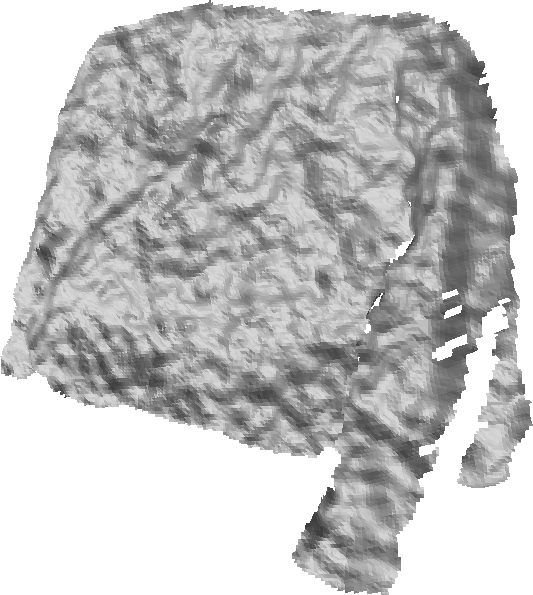}&
  \includegraphics[width=\mywidth]{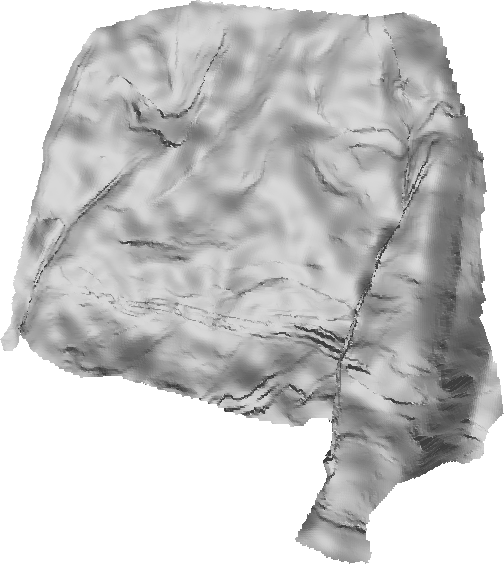}&
  \includegraphics[width=\mywidth]{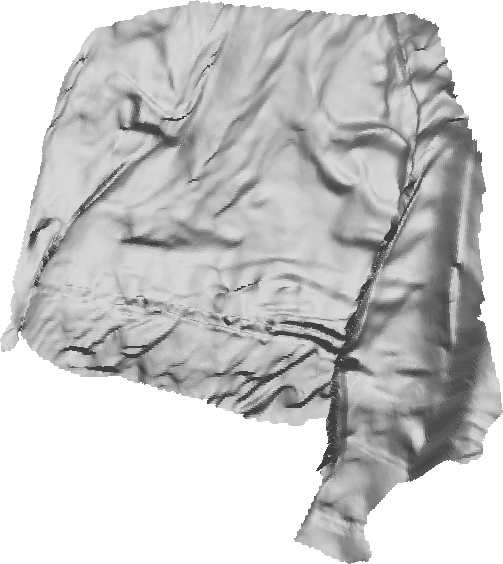}&
  \includegraphics[width=\mywidth]{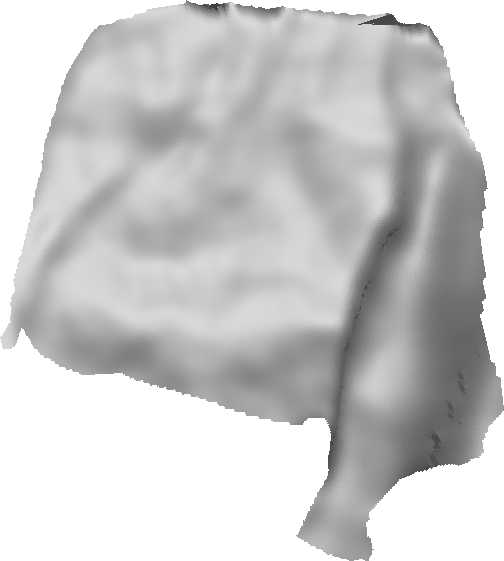}&
  \includegraphics[width=\mywidth]{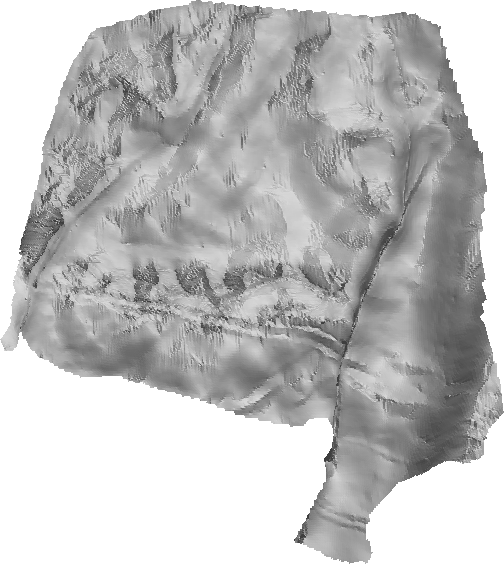}&
  \includegraphics[width=\mywidth]{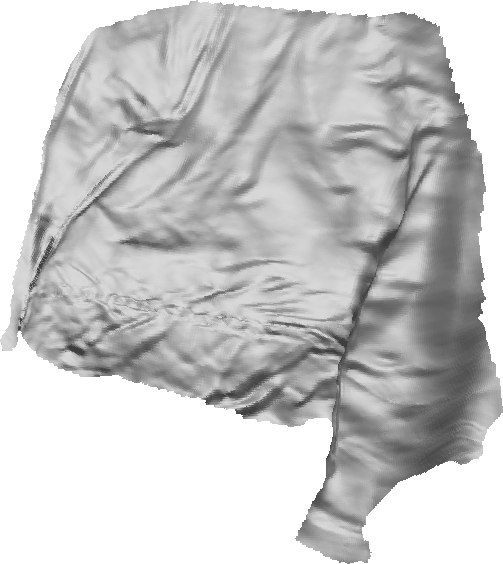}\\
  \rotatebox{90}{Basecap}&
  \includegraphics[width=\mywidths]{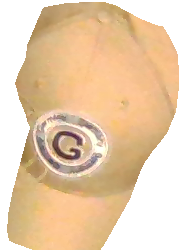}&
  \includegraphics[width=\mywidthlr]{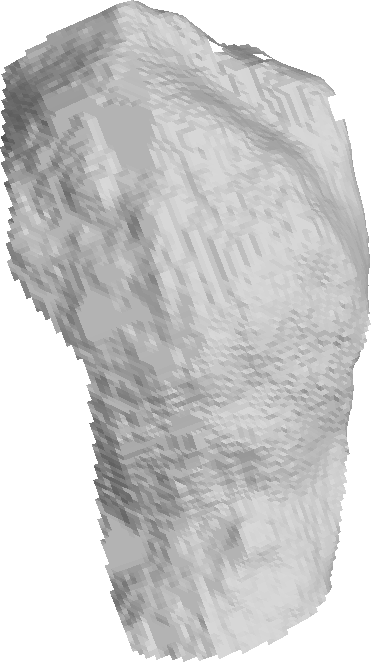}&
  \includegraphics[width=\mywidths]{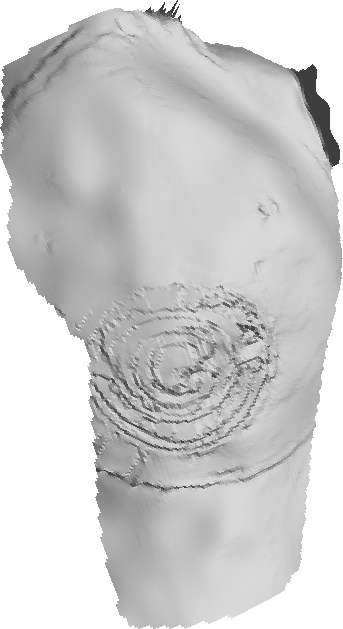}&
  \includegraphics[width=\mywidths]{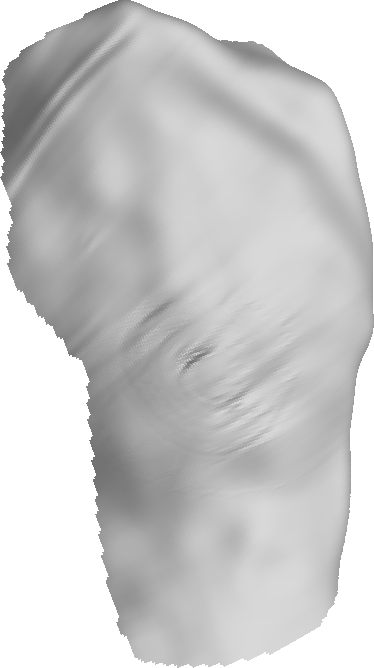}&
  \includegraphics[width=\mywidths]{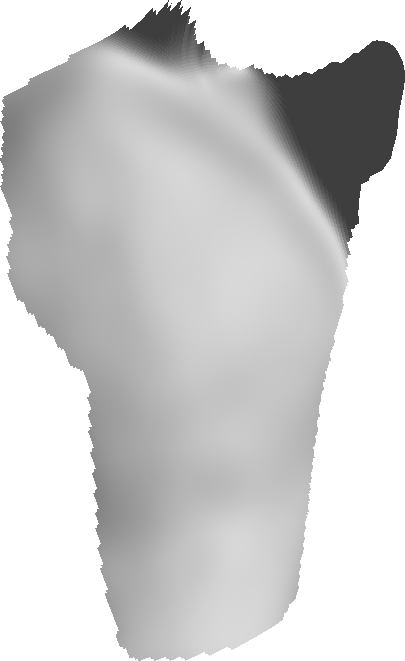}&
  \includegraphics[width=\mywidths]{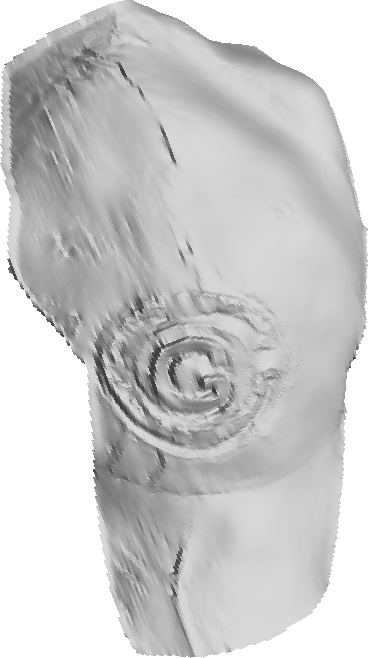}&
  \includegraphics[width=\mywidths]{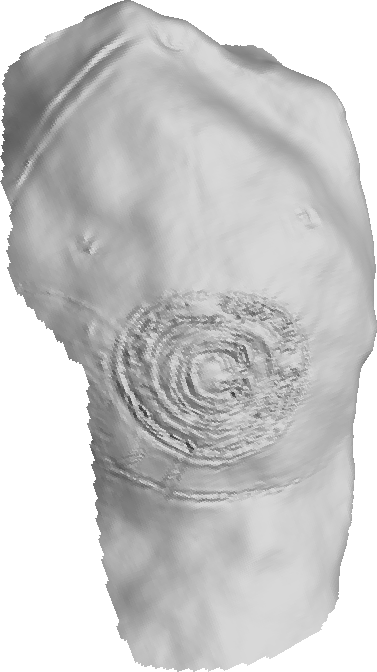}\\
  \rotatebox{90}{Hat}&
  \includegraphics[width=\mywidth]{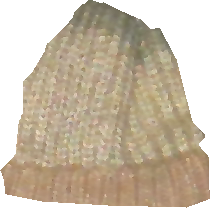}&
  \includegraphics[width=\mywidthlr]{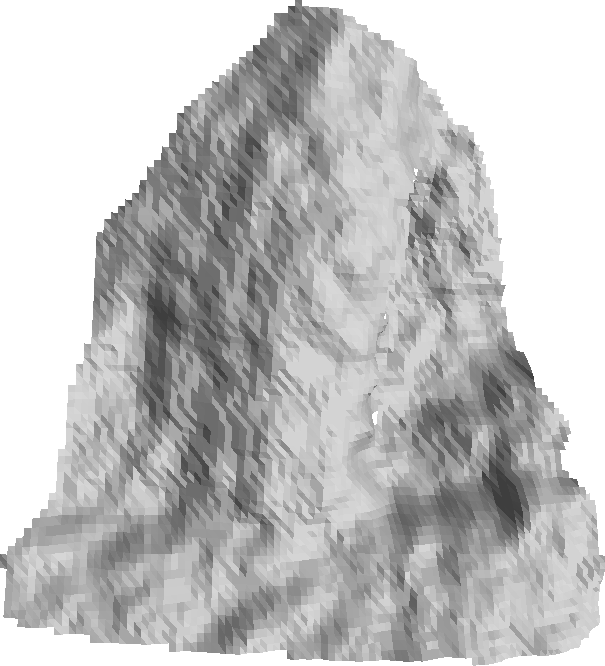}&
  \includegraphics[width=\mywidth]{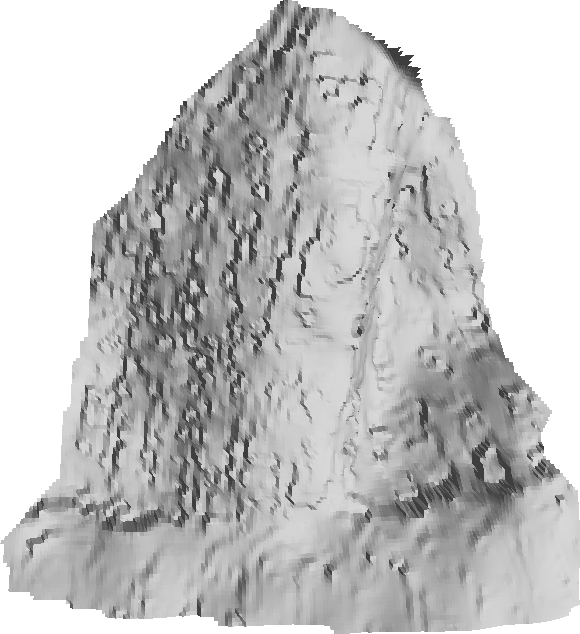}&
  \includegraphics[width=\mywidth]{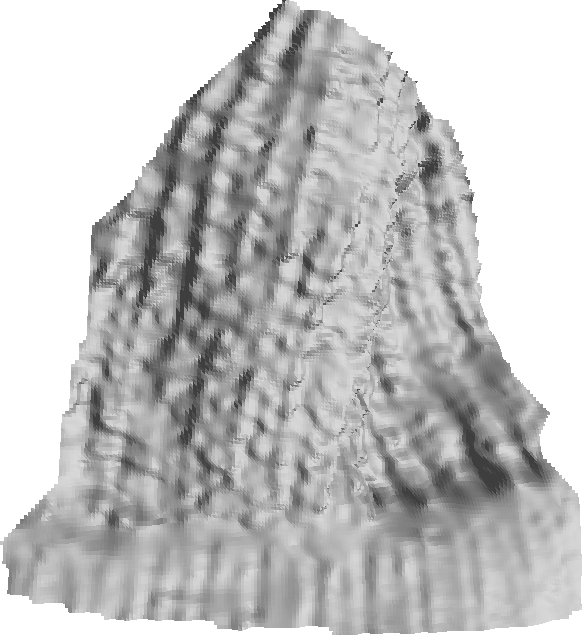}&
  \includegraphics[width=\mywidth]{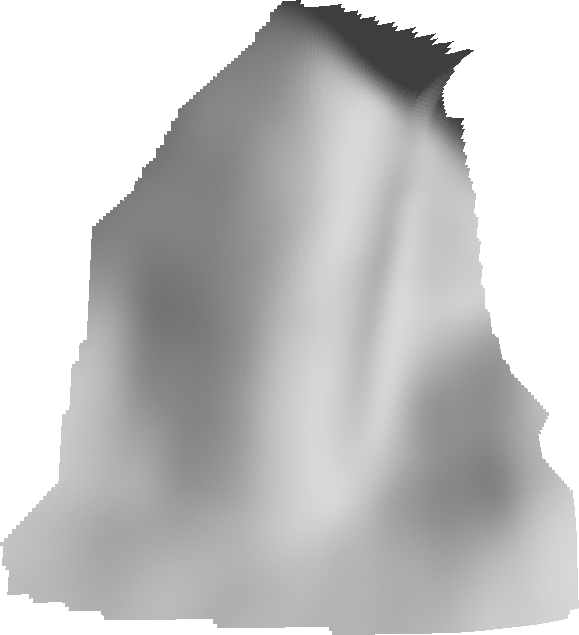}&
  \includegraphics[width=\mywidth]{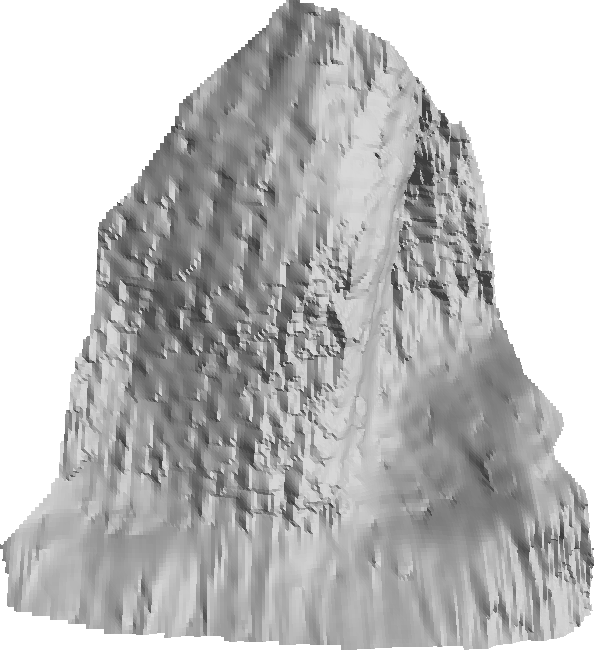}&
  \includegraphics[width=\mywidth]{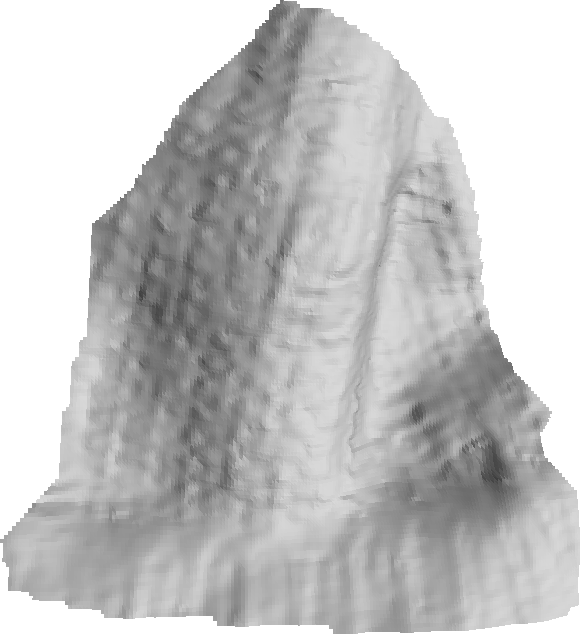}\\
  \rotatebox{90}{Pillow}&
  \includegraphics[width=\mywidth]{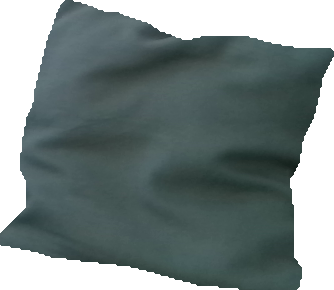}&
  \includegraphics[width=\mywidthlr]{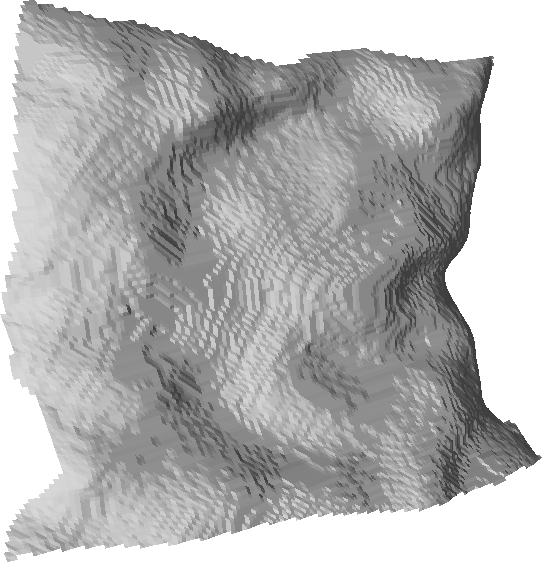}&
  \includegraphics[width=\mywidth]{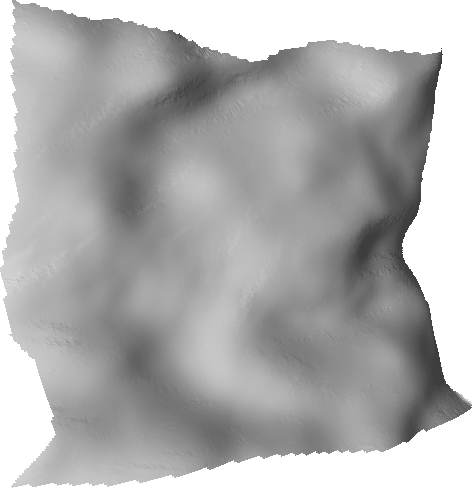}&
  \includegraphics[width=\mywidth]{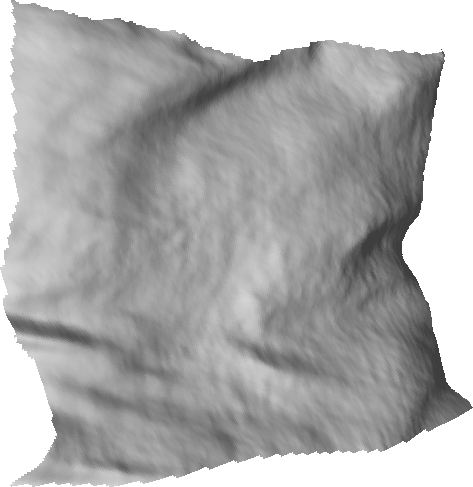}&
  \includegraphics[width=\mywidth]{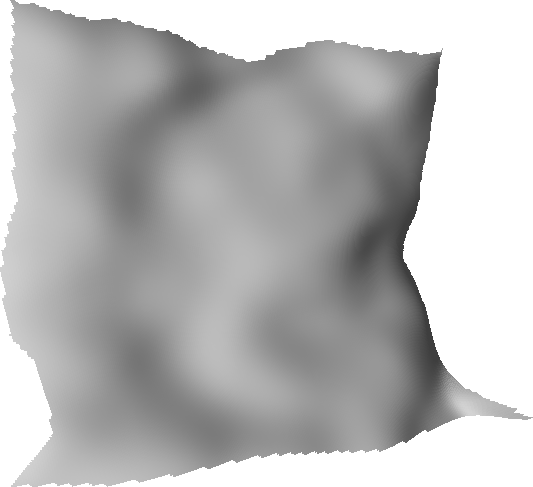}&
  \includegraphics[width=\mywidth]{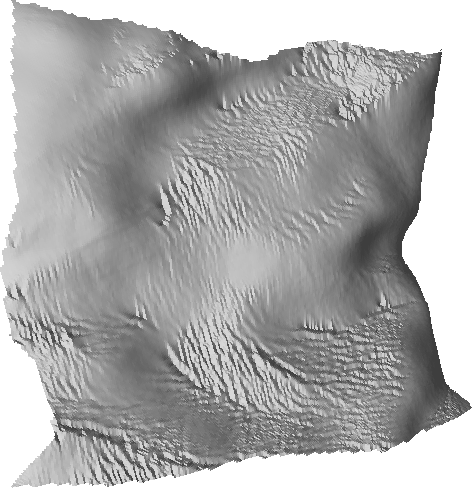}&
  \includegraphics[width=\mywidth]{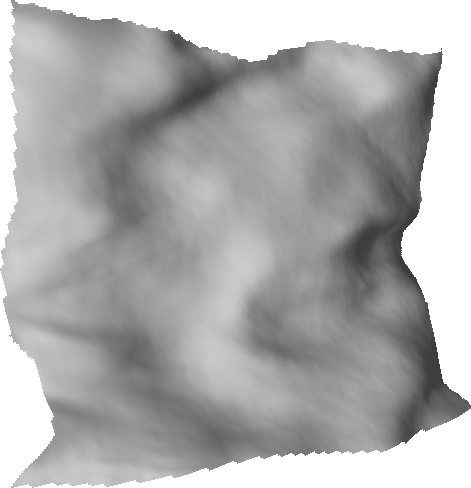}
  \end{tabular}
  \caption{Results of state-of-the-art methods and our approach on challenging synthetic (Joyful Yell and Stanford Bunny) and real-world (rest) datasets.
  Our approach is the only one giving satisfactory results across all datasets.}
  \label{fig:comparison}
\end{figure*}

Real-world data is captured using an Intel RealSense D415 RGB-D camera with attached LED light source, cf. Figure~\ref{fig:setup}. 
The resolution of RGB and depth is $1280\times720$ and $640\times360$, respectively.
We captured $n=20$ images of each object shown in Figure~\ref{fig:comparison} by moving the camera at a distance of $0.5$--$2$m under low ambient light such that the LED illumination dominates.
Although automatic approaches exist~\cite{haefner20193dv}, we manually segment each image, and the first frame defines the reference image $\mI_0$ and depth $\zz$.
No further calibration is needed while $f$ and $\pc$ can be accessed using Intels RealSense camera software~\cite{librealsense}.

The qualitative results can be seen in Figure~\ref{fig:comparison}.
While too smooth depth estimates are recovered for~\cite{li2016}, better results under the appearance of some remaining noise (Joyful Yell and Hat) are achieved by~\cite{liu2017}.
The importance of joint estimation of camera motion and geometry, albedo and lighting becomes apparent when comparing our results to the ones of~\cite{peng2017}.
Although, the results of~\cite{peng2017} are satisfactory on the Jacket and Hat dataset, the approach is not able to recover the fine details of the stitching on the Basecap.
These fine details are recovered by~\cite{haefner2018cvpr}, but artifacts are visible on Jacket and Pillow. Our approach gives satisfactory results across all datasets, showing that multi-view uncalibrated photometric stereo is indeed helpful to recover high resolution fine detailed depth.

\section{Conclusion}\label{sec6}
We presented a novel approach to depth super-resolution using multi-view uncalibrated photometric stereo.
We use a simple setup by attaching an LED light source to an RGB-D camera.
Different illumination of the scene is achieved by moving the camera capturing an object from different viewpoints.
No calibration on lighting and camera motion is necessary, due to our end-to-end variational approach which jointly optimizes over camera motion, geometry, reflectance and lighting.
We demonstrated on multiple challenging synthetic and real-world datasets the superiority of our method compared to other state-of-the-art approaches.
\\
In the future, we aim at reducing the error if too many images are used.
Furthermore, modelling the LED as a point light source, instead of spherical harmonic approximation or use all available depth measurements from the RGB-D sensor, instead of just the reference depth seem to be promising directions in order to further improve geometry estimation. 

{\small
\bibliographystyle{ieee}
\bibliography{references}

\begin{thebibliography}{10}\itemsep=-1pt

\bibitem{bendansie2015}
{The Joyful Yell}, 2015.
\newblock \url{https://www.thingiverse.com/thing:897412}.

\bibitem{librealsense}
{Intel RealSense SDK 2.0}, 2019.
\newblock \url{https://github.com/IntelRealSense/librealsense}.

\bibitem{abrams2012}
A.~Abrams, C.~Hawley, and R.~Pless.
\newblock Heliometric stereo: Shape from sun position.
\newblock In {\em European Conference on Computer Vision}, pages 357--370,
  2012.

\bibitem{ackermann2012}
J.~Ackermann, F.~Langguth, S.~Fuhrmann, and M.~Goesele.
\newblock Photometric stereo for outdoor webcams.
\newblock In {\em 2012 IEEE Conference on Computer Vision and Pattern
  Recognition}, pages 262--269, 2012.

\bibitem{ackermann2014}
J.~{Ackermann}, F.~{Langguth}, S.~{Fuhrmann}, A.~{Kuijper}, and M.~{Goesele}.
\newblock Multi-view photometric stereo by example.
\newblock In {\em 2014 International Conference on 3D Vision}, pages 259--266,
  2014.

\bibitem{basri2007}
R.~Basri, D.~Jacobs, and I.~Kemelmacher.
\newblock Photometric stereo with general, unknown lighting.
\newblock {\em International Journal of Computer Vision}, 72(3):239--257, 2007.

\bibitem{basri2003}
R.~{Basri} and D.~W. {Jacobs}.
\newblock Lambertian reflectance and linear subspaces.
\newblock {\em IEEE Transactions on Pattern Analysis and Machine Intelligence},
  25(2):218--233, 2003.

\bibitem{belhumeur1999}
P.~N. Belhumeur, D.~J. Kriegman, and A.~L. Yuille.
\newblock The bas-relief ambiguity.
\newblock {\em International Journal of Computer Vision}, 35(1):33--44, 1999.

\bibitem{brahimi2019}
M.~Brahimi, Y.~Quéau, B.~Haefner, and D.~Cremers.
\newblock On well-posedness of uncalibrated photometric stereo under general
  lighting.
\newblock {\em arXiv preprint arXiv:1911.07268}, 2019.

\bibitem{bylow2019}
E.~Bylow, R.~Maier, F.~Kahl, and C.~Olsson.
\newblock Combining depth fusion and photometric stereo for fine-detailed 3d
  models.
\newblock In {\em Scandinavian Conference on Image Analysis}, pages 261--274,
  2019.

\bibitem{cristani2004}
M.~Cristani, D.~S. Cheng, V.~Murino, and D.~Pannullo.
\newblock Distilling information with super-resolution for video surveillance.
\newblock In {\em Proceedings of the ACM 2Nd International Workshop on Video
  Surveillance \& Sensor Networks}, pages 2--11, 2004.

\bibitem{dong2014}
C.~Dong, C.~C. Loy, K.~He, and X.~Tang.
\newblock Learning a deep convolutional network for image super-resolution.
\newblock In {\em European conference on computer vision}, pages 184--199,
  2014.

\bibitem{dong2015}
C.~Dong, C.~C. Loy, K.~He, and X.~Tang.
\newblock Image super-resolution using deep convolutional networks.
\newblock {\em IEEE transactions on Pattern Analysis and Machine Intelligence},
  38(2):295--307, 2015.

\bibitem{farsiu2004}
S.~Farsiu, D.~Robinson, M.~Elad, and P.~Milanfar.
\newblock Advances and challenges in super-resolution.
\newblock {\em International Journal of Imaging Systems and Technology},
  14(2):47--57, 2004.

\bibitem{ferstl2013}
D.~Ferstl, C.~Reinbacher, R.~Ranftl, M.~R{\"u}ther, and H.~Bischof.
\newblock Image guided depth upsampling using anisotropic total generalized
  variation.
\newblock In {\em Proceedings of the IEEE International Conference on Computer
  Vision}, pages 993--1000, 2013.

\bibitem{goldluecke2014}
B.~Goldluecke, M.~Aubry, K.~Kolev, and D.~Cremers.
\newblock A super-resolution framework for high-accuracy multiview
  reconstruction.
\newblock {\em International Journal of Computer Vision}, 106:172--191, 2014.

\bibitem{graber2015}
G.~Graber, J.~Balzer, S.~Soatto, and T.~Pock.
\newblock Efficient minimal-surface regularization of perspective depth maps in
  variational stereo.
\newblock In {\em Proceedings of the IEEE Conference on Computer Vision and
  Pattern Recognition}, pages 511--520, 2015.

\bibitem{greenspan2008}
H.~Greenspan.
\newblock Super-resolution in medical imaging.
\newblock {\em The Computer Journal}, 52(1):43--63, 2008.

\bibitem{haefner2019tpami}
B.~{Haefner}, S.~{Peng}, A.~{Verma}, Y.~{Quéau}, and D.~{Cremers}.
\newblock Photometric depth super-resolution.
\newblock {\em IEEE Transactions on Pattern Analysis and Machine Intelligence},
  pages 1--1, 2019.

\bibitem{haefner20193dv}
B.~Haefner, Y.~Quéau, and D.~Cremers.
\newblock Photometric segmentation: Simultaneous photometric stereo and
  masking.
\newblock In {\em International Conference on 3D Vision}, 2019.

\bibitem{haefner2018cvpr}
B.~Haefner, Y.~Quéau, T.~Möllenhoff, and D.~Cremers.
\newblock Fight ill-posedness with ill-posedness: Single-shot variational depth
  super-resolution from shading.
\newblock In {\em IEEE Conference on Computer Vision and Pattern Recognition},
  2018.

\bibitem{haefner2019iccv}
B.~Haefner, Z.~Ye, M.~Gao, T.~Wu, Y.~Quéau, and D.~Cremers.
\newblock Variational uncalibrated photometric stereo under general lighting.
\newblock In {\em International Conference on Computer Vision}, 2019.

\bibitem{hernandez2008}
C.~Hernandez, G.~Vogiatzis, and R.~Cipolla.
\newblock Multiview photometric stereo.
\newblock {\em IEEE Transactions on Pattern Analysis and Machine Intelligence},
  30(3):548--554, 2008.

\bibitem{higo2009}
T.~Higo, Y.~Matsushita, N.~Joshi, and K.~Ikeuchi.
\newblock A hand-held photometric stereo camera for 3-d modeling.
\newblock In {\em 2009 IEEE International Conference on Computer Vision}, pages
  1234--1241, 2009.

\bibitem{horn1989}
B.~K.~P. Horn and M.~J. Brooks, editors.
\newblock {\em Shape from Shading}.
\newblock MIT Press, 1989.

\bibitem{huhle2010}
B.~Huhle, T.~Schairer, P.~Jenke, and W.~Stra{\ss}er.
\newblock Fusion of range and color images for denoising and resolution
  enhancement with a non-local filter.
\newblock {\em Computer Vision and Image Understanding}, 114(12):1336--1345,
  2010.

\bibitem{takwai2016}
T.-W. Hui, C.~C. Loy, and X.~Tang.
\newblock Depth map super-resolution by deep multi-scale guidance.
\newblock In B.~Leibe, J.~Matas, N.~Sebe, and M.~Welling, editors, {\em
  European Conference on Computer Vision}, pages 353--369. Springer
  International Publishing, 2016.

\bibitem{jung2019}
J.~Jung, J.-Y. Lee, and I.~S. Kweon.
\newblock One-day outdoor photometric stereo using skylight estimation.
\newblock {\em International Journal of Computer Vision}, pages 1--17, 2019.

\bibitem{kerl2013}
C.~Kerl, J.~Sturm, and D.~Cremers.
\newblock Robust odometry estimation for rgb-d cameras.
\newblock In {\em International Conference on Robotics and Automation}, 2013.

\bibitem{khoshelham2012}
K.~Khoshelham and S.~O. Elberink.
\newblock Accuracy and resolution of kinect depth data for indoor mapping
  applications.
\newblock {\em Sensors}, 12(2):1437--1454, 2012.

\bibitem{kiechle2013}
M.~Kiechle, S.~Hawe, and M.~Kleinsteuber.
\newblock A joint intensity and depth co-sparse analysis model for depth map
  super-resolution.
\newblock In {\em Proceedings of the IEEE International Conference on Computer
  Vision}, pages 1545--1552, 2013.

\bibitem{ledig2017}
C.~Ledig, L.~Theis, F.~Husz{\'a}r, J.~Caballero, A.~Cunningham, A.~Acosta,
  A.~Aitken, A.~Tejani, J.~Totz, Z.~Wang, et~al.
\newblock Photo-realistic single image super-resolution using a generative
  adversarial network.
\newblock In {\em Proceedings of the IEEE Conference on Computer Vision and
  Pattern Recognition}, pages 4681--4690, 2017.

\bibitem{li2016}
Y.~Li, D.~Min, M.~N. Do, and J.~Lu.
\newblock Fast guided global interpolation for depth and motion.
\newblock In {\em European Conference on Computer Vision}, 2016.

\bibitem{li2012}
Y.~Li, T.~Xue, L.~Sun, and J.~Liu.
\newblock Joint example-based depth map super-resolution.
\newblock In {\em 2012 IEEE International Conference on Multimedia and Expo},
  pages 152--157, 2012.

\bibitem{liu2017}
W.~{Liu}, X.~{Chen}, J.~{Yang}, and Q.~{Wu}.
\newblock Robust color guided depth map restoration.
\newblock {\em IEEE Transactions on Image Processing}, 26(1):315--327, 2017.

\bibitem{logothetis2018}
F.~Logothetis, R.~Mecca, and R.~Cipolla.
\newblock A differential volumetric approach to multi-view photometric stereo.
\newblock {\em arXiv preprint arXiv:1811.01984}, 2018.

\bibitem{lu2011}
J.~Lu, D.~Min, R.~S. Pahwa, and M.~N. Do.
\newblock A revisit to mrf-based depth map super-resolution and enhancement.
\newblock In {\em 2011 IEEE International Conference on Acoustics, Speech and
  Signal Processing}, pages 985--988, 2011.

\bibitem{lu2013}
Z.~Lu, Y.-W. Tai, F.~Deng, M.~Ben-Ezra, and M.~S. Brown.
\newblock A 3d imaging framework based on high-resolution photometric-stereo
  and low-resolution depth.
\newblock {\em International Journal of Computer Vision}, 102(1-3):18--32,
  2013.

\bibitem{ma2012}
Y.~Ma, S.~Soatto, J.~Kosecka, and S.~S. Sastry.
\newblock {\em An invitation to 3-d vision: from images to geometric models},
  volume~26.
\newblock Springer Science \& Business Media, 2012.

\bibitem{maier2015}
R.~Maier, J.~St{\"u}ckler, and D.~Cremers.
\newblock Super-resolution keyframe fusion for 3d modeling with high-quality
  textures.
\newblock In {\em 2015 International Conference on 3D Vision}, pages 536--544,
  2015.

\bibitem{mumford1994}
D.~Mumford.
\newblock Bayesian rationale for the variational formulation.
\newblock In {\em Geometry-driven Diffusion in Computer Vision}, pages
  135--146. Springer, 1994.

\bibitem{or2015}
R.~Or-El, G.~Rosman, A.~Wetzler, R.~Kimmel, and A.~M. Bruckstein.
\newblock Rgbd-fusion: Real-time high precision depth recovery.
\newblock In {\em Proceedings of the IEEE Conference on Computer Vision and
  Pattern Recognition}, pages 5407--5416, 2015.

\bibitem{park2014}
J.~Park, H.~Kim, Y.-W. Tai, M.~S. Brown, and I.~S. Kweon.
\newblock High-quality depth map upsampling and completion for rgb-d cameras.
\newblock {\em IEEE Transactions on Image Processing}, 23(12):5559--5572, 2014.

\bibitem{peng2017}
S.~Peng, B.~Haefner, Y.~Quéau, and D.~Cremers.
\newblock Depth super-resolution meets uncalibrated photometric stereo.
\newblock In {\em {International Conference on Computer Vision Workshops}},
  2017.

\bibitem{queau2017jmiv}
Y.~Qu{\'{e}}au, B.~Durix, T.~Wu, D.~Cremers, F.~Lauze, and J.-D. Durou.
\newblock Led-based photometric stereo: Modeling, calibration and numerical
  solution.
\newblock {\em Journal of Mathematical Imaging and Vision}, 07 2017.

\bibitem{queau2018jmiv}
Y.~Qu{\'e}au, J.-D. Durou, and J.-F. Aujol.
\newblock Normal integration: a survey.
\newblock {\em Journal of Mathematical Imaging and Vision}, 60(4):576--593,
  2018.

\bibitem{queau2017cvpr}
Y.~Quéau, T.~Wu, F.~Lauze, J.-D. Durou, and D.~Cremers.
\newblock A non-convex variational approach to photometric stereo under
  inaccurate lighting.
\newblock In {\em IEEE Conference on Computer Vision and Pattern Recognition},
  2017.

\bibitem{riegler2016}
G.~Riegler, D.~Ferstl, M.~R{\"u}ther, and H.~Bischof.
\newblock A deep primal-dual network for guided depth super-resolution.
\newblock {\em arXiv preprint arXiv:1607.08569}, 2016.

\bibitem{gernot2016}
G.~Riegler, M.~R{\"u}ther, and H.~Bischof.
\newblock Atgv-net: Accurate depth super-resolution.
\newblock In B.~Leibe, J.~Matas, N.~Sebe, and M.~Welling, editors, {\em
  European Conference on Computer Vision}, pages 268--284, 2016.

\bibitem{simakov2003}
D.~Simakov, D.~Frolova, and R.~Basri.
\newblock Dense shape reconstruction of a moving object under arbitrary,
  unknown lighting.
\newblock In {\em Proceedings of the IEEE International Conference on Computer
  Vision}, pages 1202--1209, 2003.

\bibitem{xibin2017}
X.~Song, Y.~Dai, and X.~Qin.
\newblock Deep depth super-resolution: Learning depth super-resolution using
  deep convolutional neural network.
\newblock In S.-H. Lai, V.~Lepetit, K.~Nishino, and Y.~Sato, editors, {\em
  Asian Conference on Computer Vision}, pages 360--376, 2017.

\bibitem{treuille2004}
A.~Treuille, A.~Hertzmann, and S.~M. Seitz.
\newblock Example-based stereo with general brdfs.
\newblock In {\em European Conference on Computer Vision}, pages 457--469,
  2004.

\bibitem{turk1994}
G.~Turk and M.~Levoy.
\newblock Zippered polygon meshes from range images.
\newblock In {\em Proceedings of the 21st annual Conference on Computer
  Graphics and Interactive Techniques}, pages 311--318, 1994.

\bibitem{ullman1979}
S.~Ullman.
\newblock The interpretation of structure from motion.
\newblock {\em Proceedings of the Royal Society of London. Series B. Biological
  Sciences}, 203(1153):405--426, 1979.

\bibitem{unger2010}
M.~Unger, T.~Pock, M.~Werlberger, and H.~Bischof.
\newblock A convex approach for variational super-resolution.
\newblock In {\em Proceedings of the 32nd {DAGM} Conference on Pattern
  Recognition}, pages 313--322, 2010.

\bibitem{woodham1980}
R.~J. Woodham.
\newblock Photometric method for determining surface orientation from multiple
  images.
\newblock {\em Optical Engineering}, 19(1):139 -- 144, 1980.

\bibitem{yang2008}
J.~Yang, J.~Wright, T.~Huang, and Y.~Ma.
\newblock Image super-resolution as sparse representation of raw image patches.
\newblock In {\em 2008 IEEE Conference on Computer Vision and Pattern
  Recognition}, pages 1--8, 2008.

\bibitem{yang2010}
J.~Yang, J.~Wright, T.~S. Huang, and Y.~Ma.
\newblock Image super-resolution via sparse representation.
\newblock {\em IEEE Transactions on Image Processing}, 19(11):2861--2873, 2010.

\bibitem{yang2007}
Q.~Yang, R.~Yang, J.~Davis, and D.~Nist{\'e}r.
\newblock Spatial-depth super resolution for range images.
\newblock {\em IEEE Conference on Computer Vision and Pattern Recognition},
  pages 1--8, 2007.

\bibitem{zhang2003}
L.~Zhang et~al.
\newblock Shape and motion under varying illumination: Unifying structure from
  motion, photometric stereo, and multiview stereo.
\newblock In {\em In Proceedings of the IEEE International Conference on
  Computer Vision}, pages 618--625, 2003.

\bibitem{zhou2013}
Z.~Zhou, Z.~Wu, and P.~Tan.
\newblock Multi-view photometric stereo with spatially varying isotropic
  materials.
\newblock In {\em Proceedings of the IEEE Conference on Computer Vision and
  Pattern Recognition}, pages 1482--1489, 2013.

\bibitem{zuo2017}
X.~Zuo, S.~Wang, J.~Zheng, and R.~Yang.
\newblock Detailed surface geometry and albedo recovery from rgb-d video under
  natural illumination.
\newblock In {\em Proceedings of the IEEE International Conference on Computer
  Vision}, pages 3133--3142, 2017.

\end{thebibliography}
}

\end{document}